%% file: main.tex
\PassOptionsToPackage{table}{xcolor}
\documentclass[10pt,letterpaper]{article}
\usepackage[T1]{fontenc} % Standard text font encoding.
\usepackage{emoreward_arxiv}

\usepackage{amsmath,amssymb,amsfonts}
\usepackage{microtype}
\usepackage{graphicx}
\usepackage{fontawesome5}
\usepackage{booktabs}
\usepackage{xcolor}
\usepackage{multirow}
\usepackage{algorithm}
\usepackage{algpseudocode}
\usepackage{url}
\usepackage{hyperref}
\newcommand{\xdocref}[1]{\ref{#1}}
\hypersetup{colorlinks=true,linkcolor=hunyuanblue,citecolor=hunyuanblue,urlcolor=hunyuanblue,
  pdftitle={EmoReward: Self-Evolving Reward Calibration for Emotional Video Generation RL},
  pdfauthor={Mingzhe Zheng, Yue Wu, Jing Tan, Chang Zou, Dengyang Jiang, Zeyue Tian, Xuanhua He, Ryan Xu, Yunyang Ge, Sihui Ji, Yunlong Lin, Kaihang Pan, Hongru Yan, Sixiang Chen, Dianyi Wang, Julian Zhang, Weijie Kong, Changlin Li, Lanqing Guo, Zhao Zhong, Qifeng Chen, Harry Yang}}

\graphicspath{{figures/}{figures/Figure_GPT_0709/}}

\title{EmoReward: Self-Evolving Reward Calibration for Emotional Video Generation RL}

\author{
Mingzhe Zheng\textsuperscript{1,2,*} \quad Yue Wu\textsuperscript{2,\ensuremath{\ddagger}} \quad Jing Tan\textsuperscript{2} \quad Chang Zou\textsuperscript{2} \quad Dengyang Jiang\textsuperscript{1} \quad Zeyue Tian\textsuperscript{1}\\[2pt]
Xuanhua He\textsuperscript{1} \quad Ryan Xu\textsuperscript{2} \quad Yunyang Ge\textsuperscript{2} \quad Sihui Ji\textsuperscript{2} \quad Yunlong Lin\textsuperscript{2} \quad Kaihang Pan\textsuperscript{2}\\[2pt]
Hongru Yan\textsuperscript{2} \quad Sixiang Chen\textsuperscript{2} \quad Dianyi Wang\textsuperscript{3} \quad Julian Zhang\textsuperscript{4} \quad Weijie Kong\textsuperscript{2}\\[2pt]
Changlin Li\textsuperscript{2} \quad Lanqing Guo\textsuperscript{2} \quad Zhao Zhong\textsuperscript{2} \quad Qifeng Chen\textsuperscript{1,\ensuremath{\dagger}} \quad Harry Yang\textsuperscript{1,\ensuremath{\dagger}}\\[7pt]
{\normalfont\fontsize{9}{11}\selectfont
\textsuperscript{*}Work done during internship at Tencent Hunyuan\\[2pt]
\textsuperscript{\ensuremath{\ddagger}}Project leader\qquad
\textsuperscript{\ensuremath{\dagger}}Corresponding author}\\[6pt]
{\fontsize{9}{11}\selectfont
\textsuperscript{1}The Hong Kong University of Science and Technology\quad
\textsuperscript{2}Tencent Hunyuan\\[2pt]
\textsuperscript{3}Fudan University\quad
\textsuperscript{4}University of Waterloo}
}

\begin{document}
\maketitle

\input{sections/00_abstract}
\clearpage
\input{sections/01_introduction}
\input{sections/02_related_work}
\input{sections/03_method}

\begingroup
\setlength{\parskip}{3pt plus 1pt minus 1pt}
\setlength{\textfloatsep}{12pt plus 2pt minus 2pt}
\setlength{\floatsep}{10pt plus 2pt minus 2pt}
\input{sections/04_experiment}
\input{sections/05_conclusion}
\label{main_end}
\endgroup
\clearpage
\input{sections/06_llm_usage}

\bibliographystyle{plainnat}
\bibliography{references}

\clearpage
\appendix
\makeatletter\setlength{\@fptop}{0pt}\makeatother
\input{sections/10_appendix}

\end{document}

%% file: sections/00_abstract.tex
\begin{abstract}
Group Relative Policy Optimization (GRPO) has advanced generative model post-training in verifiable domains such as code and mathematics, but evaluating video quality, especially emotional expression, often depends on subjective aesthetics and has no uniquely correct target.
Existing approaches typically rely on fixed rewards for visual quality, motion quality, and text alignment that poorly capture emotional intent and can invite reward hacking, while reinforcement learning from human feedback (RLHF) is difficult to scale and subject to individual aesthetic differences that complicate the calibration of facial, body motion, and chromatic cues across narrative contexts.
We introduce \textbf{EmoReward}, a new self-evolving training mechanism that distills evidence reliability from reward traces collected during an actual training run to adaptively guide multi-reward weight allocation in subsequent training by determining which emotional evidence to trust and how much to weight it.
EmoReward first plans specific grounding cues from the emotional prompt to guide evaluation, then uses heterogeneous signal primitives to score the generated video during the act stage. \emph{Reflective Reward Gating} calibrates their contributions through instance-level gates for phase, face confidence, conflict, and subtlety, then renormalizes weights over the active set to produce the sample's fused reward.
\emph{RewardMemory} then estimates each primitive's reliability from rollout scores and outcomes within each emotional regime in the finished training run, then fixes bounded multipliers that calibrate fusion weights in the next training run without updating reward models.
On HunyuanVideo~1.5, EmoReward improves held-out emotion accuracy and ranking across three RL algorithms.
Ablations and transfer experiments on SD~3.5 and FLUX further support adaptive reward calibration for visual generation.
\par\vspace{9pt}
{\sffamily\small
\noindent\makebox[1.5em][l]{\textcolor{hunyuanblue}{\faGlobe}}\textbf{Project Page: }\href{https://emoreward-project.github.io/}{\texttt{https://emoreward-project.github.io/}}\par
\vspace{4pt}
\noindent\makebox[1.5em][l]{\faGithub}\textbf{GitHub Repo: }\href{https://github.com/Tencent-Hunyuan/EmoReward}{\texttt{https://github.com/Tencent-Hunyuan/EmoReward}}\par
}
\end{abstract}

%% file: sections/01_introduction.tex
\section{Introduction}
\label{sec:intro}

Reinforcement learning (RL), especially Group Relative Policy Optimization (GRPO), has advanced model post-training in verifiable domains such as code and mathematics, where correctness provides a direct learning signal \citep{shao2024deepseekmath,guo2025deepseekr1,kimiteam2025kimi,yu2025dapo}.
In contrast, evaluating video quality, especially emotional expression, often depends on subjective aesthetics and has no uniquely correct target.
Visual post-training adapts diffusion and flow-based generation models through RL and preference objectives \citep{black2023ddpo,fan2023dpok,prabhudesai2023alignprop,wallace2023diffusiondpo,xue2025dancegrpo,liu2025flowgrpo,zheng2026manifold}. 
Learned evaluators provide scores for image preference and video quality \citep{xu2023imagereward,kirstain2023pickapic,ma2025hpsv3,xu2024visionreward,he2024videoscore,liu2025videoalign}, but a visually clear, temporally stable video can match its prompt without conveying the intended emotion. 
Facial analysis, dynamic expression recognition, and multimodal emotion models supply complementary evidence \citep{baltrusaitis2018openface2,chen2024finecliper,cheng2024emotionllama,lian2025affectgpt,zhao2025r1omni}, whose reliability varies with context: facial cues require visible faces, strong motion can undermine restrained sadness, and saturated colors may reflect style rather than emotional intent. 
This raises a central question: \emph{Which emotional evidence should the reward system trust in the current video, and how can it adaptively determine how much weight to assign to that evidence?}

\begin{figure}[t]
  \centering
  \includegraphics[width=.80\textwidth]{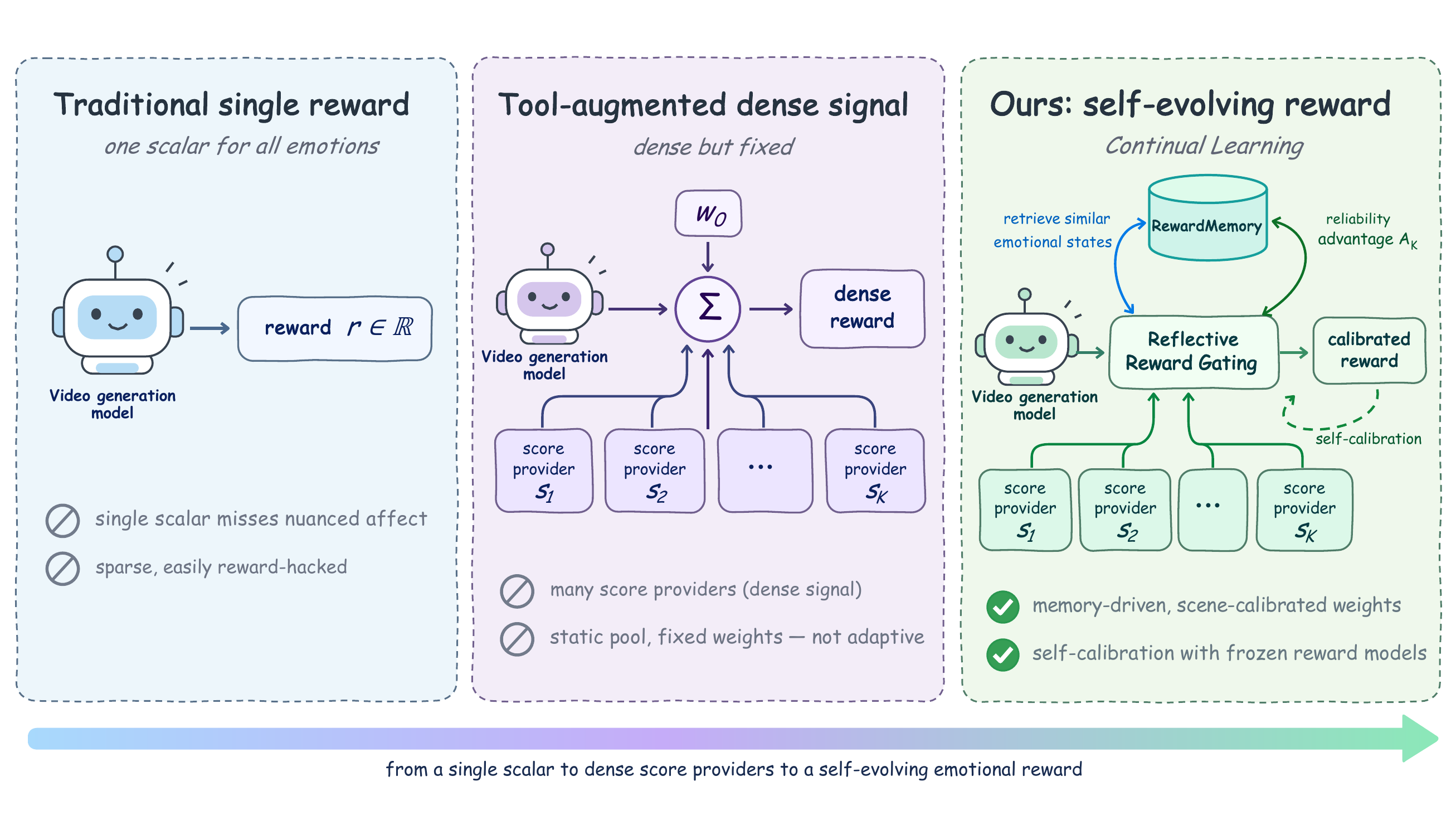}
  \caption{\textbf{From fixed scoring to contextual reward calibration.}
  The reward design progresses from a fixed evaluator to a heterogeneous provider pool and
  then to EmoReward. Reflective Reward Gating selects relevant evidence for the current
  video; RewardMemory calibrates provider trust from experience. All three stages ultimately
  return a scalar reward. The reward providers remain frozen while RL updates the video generation model.}
  \label{fig:motivate}
\end{figure}

Agentic systems coordinate interaction, refine outputs through feedback, and organize evaluation into rubrics \citep{wu2023autogen,wang2024moa,madaan2023selfrefine,shinn2023reflexion,gunjal2025rubrics}.
These capabilities motivate adaptive evaluation, yet existing video GRPO methods typically rely on fixed rewards for visual quality, motion quality, and text alignment that inadequately capture emotional intent. 
Reinforcement learning from human feedback (RLHF) and preference optimization address subjective goals \citep{christiano2017deep,stiennon2020learning,ouyang2022training,rafailov2023direct}, but human annotation is costly at scale, and aesthetic differences complicate the calibration of emotional cues across contexts. 
Fixed fusion can mix informative, missing, and saturated signals, allowing policy updates to exploit scoring shortcuts rather than improve emotional fidelity. 
This motivates reward calibration at two complementary levels: assessing which cues are valid within each video, and learning from a completed training run how much those cues should influence rewards in the next run.

We introduce \textbf{EmoReward}, a \emph{self-evolving training mechanism} for emotional video-generation RL (Figure~\ref{fig:motivate}).
It distills \emph{evidence reliability} from reward--outcome traces collected during an actual, completed training run, then uses this experience to guide multi-reward weight allocation in the next run to decide which emotional evidence to trust and how much to weight it.
\emph{Plan} translates emotional intent into an EmoScript specifying affect targets, facial and body cues, and an expression window; \emph{Act} scores generated videos with heterogeneous primitives, keeping criterion scores separate from evidence validity.
Within each video, \emph{Reflective Reward Gating} calibrates these cues using the training phase, face confidence, semantic conflict, and subtlety, then renormalizes weights over active providers.
After a training run, \emph{RewardMemory} estimates provider reliability for each emotional regime from traces. 
Providers with stronger agreement are given larger weight multipliers for the next run, with limits on the size of the adjustment. 
These multipliers and all reward models stay fixed throughout that run; for each new video, reflection further downweights or excludes missing or unsuitable cues before the active scores are combined into a weighted average.
EmoReward integrates with several RL algorithms alongside general video rewards, using training experience to calibrate reward allocation without retraining the evaluators.

We evaluate EmoReward on HunyuanVideo~1.5~\citep{wu2025hunyuanvideo15} with DanceGRPO~\citep{xue2025dancegrpo}, FlowGRPO~\citep{liu2025flowgrpo}, and SAGE-GRPO~\citep{zheng2026manifold}, finding higher held-out emotion accuracy and rank scores across all three RL algorithms and higher VideoAlign Overall with FlowGRPO and SAGE-GRPO. 
Under a shared architecture and training budget, RewardMemory yields the highest reliability selectivity among the seven reward policies. 
Ablation studies and mechanism analyses examine the contributions of reflective gating, provider pools, and RewardMemory (Section~\ref{sec:exp2}). 
Transfer experiments on Stable Diffusion~3.5 and FLUX show gains in selected configurations alongside objective-dependent trade-offs, while user studies show that EmoReward receives the highest proportion of Good ratings and the lowest proportion of Bad ratings for emotional fidelity, temporal coherence, and visual quality.

Our contributions are threefold.
\begin{itemize}
  \setlength{\itemsep}{1pt}
  \interlinepenalty=10000 % Keep each contribution paragraph on one page.
  \item We introduce \textbf{EmoReward}, the first self-evolving training mechanism for emotional video generation RL that calibrates reward providers across training runs without updating reward model parameters.
  \item We propose \textbf{RewardMemory}, the first memory designed for multi-reward systems in video generation. It distills evidence reliability into bounded fusion weights, enabling iterative calibration of RL reward signals across training runs.
  \item We introduce a \textbf{cross-run self-evolution paradigm} for post-training multimodal foundation models: a completed training run informs reward allocation in the next, making training experience reusable at the run level. Extensive experiments across three video RL algorithms and two image backbones, with ablations and human evaluations, validate EmoReward's effectiveness.
\end{itemize}

%% file: sections/02_related_work.tex
\section{Related Work}
\label{sec:related}

\paragraph{From verifiable RL to agentic feedback for non-verifiable rewards.}
RLHF learns subjective preferences \citep{christiano2017deep,ouyang2022training}, whereas verifiable RL exploits correctness checks \citep{guo2025deepseekr1,yu2025dapo}.
Agentic evaluators \citep{zhang2025landscape,cui2026agentic} and adaptive rubrics \citep{gunjal2025rubrics,xu2026rubricarm,zhang2026qubric} structure open-ended feedback.
Related approaches route experts \citep{ong2024routellm}, refine responses \citep{madaan2023selfrefine}, evolve rewards \citep{ma2024eureka,yuan2024selfrewarding}, or organize experience \citep{shinn2023reflexion,xu2025amem}.
JitRL derives closed-form policy updates from retrieved action advantages \citep{li2026jitrl}.
EmoReward instead transfers reward--outcome evidence across completed training runs to calibrate frozen providers, combining sample-level gating with persistent fusion weights.

% \paragraph{Agents for visual content creation.}
% Agentic visual creation organizes generation into planning, tool use, and feedback-driven refinement.
% VideoGen-of-Thought decomposes multi-shot synthesis into narrative planning, keyframe generation, and consistent shot synthesis \citep{zheng2024videogen}; related work studies script coherence and character consistency \citep{zheng2025cml}, while explicit semantic blueprints coordinate joint video--audio generation \citep{tu2026baton}.
% For images, agents learn to retrieve external knowledge and orchestrate generation tools \citep{feng2026gen,chen2026genevolve}, supported by advances in multimodal search \citep{chen2026opensearch}.
% JarvisArt learns a tool-using photo-retouching policy, while JarvisEvo jointly optimizes editing and evaluation \citep{lin2025jarvisart,lin2025jarvisevo}.
% Much of this work centers on constructing inference workflows or improving the agent policies that execute them; learned extensions can also distill visual experience or update evaluators.
% EmoReward brings self-evolution into the reward mechanism used during emotional video-generation RL: to our knowledge, it is the first to distill evidence reliability from a completed training run into bounded reward-fusion multipliers for the next run, combined with instance-level gating, without updating any reward model.
\paragraph{Agents for visual content creation.}
Agentic visual creation organizes generation into planning, tool use, and feedback-driven refinement.
Video agents coordinate multi-shot narratives, script evaluation, audiovisual planning, and learned camera control\citep{zheng2024videogen,hu2026camera,zheng2025cml,fan2026videoharness,tu2026baton, wu2026campilot}.
Image agents combine knowledge retrieval \citep{feng2026gen,chen2026opensearch}, learned creation and editing policies \citep{lai2026visioncreator,lin2025jarvisart}, and experience reuse \citep{chen2026genevolve,chen2026memogen}.
Other approaches jointly improve generation and evaluation \citep{lin2025jarvisevo,thawkar2026ask}.
EmoReward naturally achieves a self-evolving mechanism in video generation RL training, rather than building a workflow or retraining reward models.

\paragraph{RLHF for visual generation and the emotion-reward gap.}
Visual reward models capture preferences and multidimensional quality \citep{xu2023imagereward,kirstain2023pickapic,wu2023hpsv2,ma2025hpsv3,xu2024visionreward,wang2026unifiedrewardflex,liu2025videoalign}.
Optimization using rewards and preferences extends from images \citep{black2023ddpo,fan2023dpok,prabhudesai2023alignprop,wallace2023diffusiondpo} to video and flow models \citep{xue2025dancegrpo,liu2025flowgrpo,zheng2026manifold,liu2025videodpo}, with benchmarks assessing quality and compositional alignment \citep{huang2024vbench,liu2024evalcrafter,sun2025t2vcompbench}.
Related 3D, audiovisual, and objectives for human animation address complementary aspects of visual generation \citep{leng2026preference,huang2026va,tu2025stableanimator,tu2025stableavatar}.
Motion editing and transfer \citep{tu2024motioneditor,tu2024motionfollower,ma2026fastvmt}, shape-aware and consistent image editing \citep{long2025follow,ma2026group}, and pose- and identity-preserving animation \citep{tu2025stableanimator++,pang2025dreamdance,tu2026flashportrait} offer finer control over visual content.
Distillation \citep{yang2026arcflow,jiang2026d,fang2026flow} and feature reuse \citep{ma2025model} improve generation efficiency, complementing reward-based post-training.
Affect-grounded augmentation \citep{wang2026argen} and reflective action-conditioned generation \citep{yang2026spiral} further connect feedback with video synthesis.
Temporal video modeling and dynamic facial-expression recognition provide additional perceptual foundations \citep{tu2023implicit,chen2024finecliper}.
Emotion understanding supplies multimodal evidence \citep{cheng2024emotionllama,lian2025affectgpt,zhao2025r1omni} and grounded affect representations \citep{russell1980circumplex,ekman1978facs,baltrusaitis2018openface2}; EmoReward addresses how their reliability should govern reward fusion across scenes and training runs.

%% file: sections/03_method.tex
\section{Method}
\label{sec:method}

\begin{figure}[t]
 \centering
 \includegraphics[width=\textwidth]{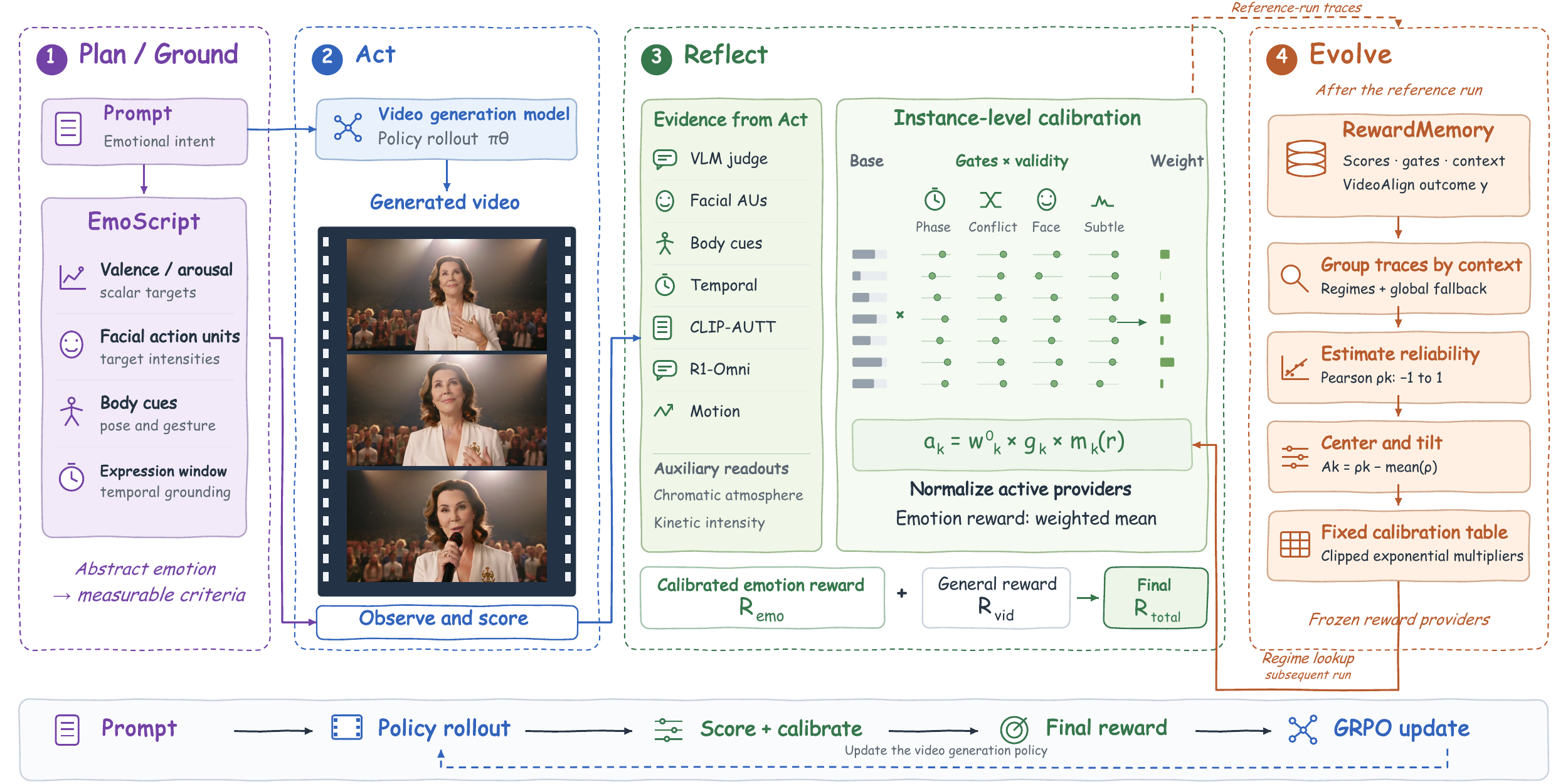}
 \caption{\textbf{EmoReward calibrates emotional evidence within videos and across training runs.}
Plan translates intent into evaluation criteria, and Act obtains provider scores and evidence validity.
Reflect combines current-video gates with memory-derived multipliers and normalizes over active providers.
Evolve estimates contextual reliability from a completed reference run to construct the fixed multipliers
used in subsequent training. The calibrated emotion reward complements the original video reward;
GRPO updates the video generation model while the providers and memory table remain fixed.}
 \label{fig:overview}
\end{figure}

\subsection{Preliminaries}
\label{sec:prelim}

\paragraph{Multi-reward optimization with GRPO.}

Let $p$ be a prompt and $x$ a generated video from a policy $\pi_\theta(x\mid p)$; a scalar
reward is $R(x,p)\in\mathbb{R}$. For a prompt $p$, GRPO samples a group
$\mathcal{G}(p)=\{x_i\}_{i=1}^{G}$ and forms standardized group advantages
\begin{equation}
\begin{aligned}
 A_i &= \frac{R(x_i,p)-\mu_{\mathcal G}}{\sigma_{\mathcal G}+\epsilon},
 &\mu_{\mathcal G}&=\tfrac{1}{G}\sum_j R(x_j,p),\\
 \sigma_{\mathcal G}^2&=\tfrac{1}{G}\sum_j(R(x_j,p)-\mu_{\mathcal G})^2.
\end{aligned}
    \label{eq:grpo_adv}
\end{equation}
which enter the clipped, KL-regularized objective in Appendix~\xdocref{app:policy_objective}. The
policy update depends on reward differences inside the group. A common additive constant
cancels exactly; positive rescaling changes only the relative effect of $\epsilon$.
A coefficient of variation measures dispersion, not policy-learning signal quality;
we also assess within-prompt ranking agreement and downstream quality.

\paragraph{Why static emotion fusion fails.}
\label{sec:static_fail}

Given reward primitives $\mathcal{T}=\{1,\ldots,K\}$ with normalized scores
$s_k(x,p)\in[0,1]$, a fixed fusion is
$R_{\mathrm{emo}}^{\mathrm{static}}=\sum_k w_k^0 s_k$ with $w_k^0\ge0,\sum_k w_k^0=1$. This
assumes a fixed trust allocation across scenes, although facial visibility and the relevance
of motion depend on the intended emotion. Writing
each primitive as a constant plus a group-varying part $s_k=c_k+\delta_k$ and splitting into
an active set $\mathcal{A}$ and a dead/saturated set $\mathcal{D}$,
\begin{equation}
    R_{\mathrm{emo}}^{\mathrm{static}}
    = \sum_{k\in\mathcal{A}}w_k^0 \delta_k
    + \sum_{k\in\mathcal{A}}w_k^0 c_k
    + \sum_{k\in\mathcal{D}}w_k^0 c_k ,
    \label{eq:static_decomp}
\end{equation}
the constant terms cancel from the centered GRPO numerator. Removing them alone need not
change the standardized advantage. Calibration matters when provider validity, errors, or
relative contributions vary across candidates, or when the emotion branch is mixed with
another reward. EmoReward addresses these cases through gates and memory-conditioned fusion.

\subsection{Overview}
\label{sec:overview}

EmoReward grounds emotional intent, scores visual evidence, and calibrates provider trust:
\begin{equation}
\begin{gathered}
 p\xrightarrow{\mathrm{Plan}} e(p),\qquad
 (x,e)\xrightarrow{\mathrm{Act}}\{s_k,u_k\},\\
 (s,u,e,x)\xrightarrow{\mathrm{Reflect}} R_{\mathrm{emo}},\qquad
 \mathcal{M}\xrightarrow{\mathrm{Evolve}}w^{\mathrm{RM}}(z).
\end{gathered}
\end{equation}
Here $e(p)$ is an EmoScript, $u_k$ denotes evidence validity, and $z$ is the emotional-context
signature. RewardMemory $\mathcal M$ stores completed reference-run traces; Evolve produces
$w^{\mathrm{RM}}(z)$ and exports the corresponding regime multipliers $m_k(r)$
(Section~\ref{sec:evolve}). Reflect combines these fixed multipliers with current-video gates.
The resulting emotion reward is mixed with a general video reward:
\begin{equation}
 R_{\mathrm{total}}(x,p)=\frac{\alpha R_{\mathrm{vid}}(x,p)+\beta R_{\mathrm{emo}}(x,p)}{\alpha+\beta},
 \label{eq:total_reward}
\end{equation}
where $R_{\mathrm{vid}}$ is the evaluation score supplied by the original reward system,
for example a general video-quality evaluator \citep{liu2025videoalign}; $\alpha,\beta\ge0$
control its mixture with the calibrated emotion reward, with $\alpha+\beta>0$.
Figure~\ref{fig:overview} connects these stages through their shared fusion interface.

\subsection{Plan and Act: Emotional Grounding}
\label{sec:plan_act}

\paragraph{Plan: prompt-conditioned criteria.}
The planner converts the prompt into an EmoScript independently of RewardMemory:
\begin{equation}
 e(p)=(v^\star,a^\star,\mathbf q^\star_{\mathrm{AU}},\mathbf b^\star,\Omega).
 \label{eq:emoscript}
\end{equation}
The implemented representation contains scalar valence and arousal targets, target facial
AU intensities, body cues, and an expression window. The memory regime is inferred later from observed context. Continuous VAD trajectories
and dominance targets are outside this implementation.
The EmoScript separates intended affect from its observable cues. For example,
restrained sadness need not imply large motion or an exaggerated face: target
intensities, body cues, and the expression window define what to assess, while
reliability is determined from the generated evidence and completed-run experience.

\paragraph{Act: observation and scoring.}
Act evaluates a video already sampled from $\pi_\theta(\cdot\mid p)$; it does not generate
or revise the video. Each provider returns
\begin{equation}
 (s_k,u_k)=\mathcal S_k(x,e(p)),\qquad s_k,u_k\in[0,1],
 \label{eq:primitive_score}
\end{equation}
where $\mathcal S_k$ is provider $k$'s scoring operator, which aggregates frame-level
measurements or evaluates the whole clip.
The seven fusion providers comprise a VLM judge, facial AUs, body cues, temporal evidence,
CLIP-AUTT, R1-Omni, and motion naturalness; chromatic atmosphere and kinetic intensity
are auxiliary readouts, giving nine observation channels. Score $s_k$ measures criterion
agreement, while $u_k$ marks usable evidence. A missing face channel's neutral default
is not evidence and is excluded through its validity mask. Auxiliary readouts do not
enter the seven-provider calibration policy. Appendix~\xdocref{app:primitives} specifies
scoring interfaces; Figure~\ref{fig:provider_pool} illustrates the toolkit studied in
Section~\ref{sec:exp3}.

\begin{algorithm}[t]
\caption{EmoReward reward computation with experience-calibrated weights}
\label{alg:emoreward}
\begin{algorithmic}[1]
\Require prompt $p$, generated video $x$, fixed table $\mathcal{T}_{\mathcal{M}}$ from a completed reference run, base weights $w^0$
\State Build EmoScript $e(p)$
\State Compute primitive scores and validity $\{s_k,u_k\}_{k=1}^{K}$ \Comment{Eq.~\ref{eq:primitive_score}}
\State Compute instance gates $\{g_k\}_{k=1}^{K}$ \Comment{Eq.~\ref{eq:gate}}
\State Form state signature $z\gets z(p,x,e,s,g)$
\State Look up regime $r(z)$ in $\mathcal{T}_{\mathcal{M}}$ \Comment{Appendix~\ref{app:retrieval}}
\If{a supported regime or global entry is available}
    \State Read fixed multipliers $m_k$, constructed by Eq.~\ref{eq:jit_weight}
\Else
    \State $m_k\gets1$ \Comment{cold start: fall back to base}
\EndIf
\State Form active weights $a_k\gets w_k^0m_kg_k$
\State Fuse active primitives with renormalization $\to R_{\mathrm{emo}}$ \Comment{Eq.~\ref{eq:active_fusion}}
\State Compute total reward $R_{\mathrm{total}}(x,p)$ \Comment{Eq.~\ref{eq:total_reward}}
\State Log the reward trace for inspection; keep $\mathcal{T}_{\mathcal{M}}$ fixed
\State \Return $R_{\mathrm{total}}(x,p)$
\end{algorithmic}
\end{algorithm}

\subsection{Reflect: Instance-Level Calibration}
\label{sec:reflect}

Reflect computes provider-specific gates from the current evidence, prompt, and training
step. The factorization is
\begin{equation}
 g_k=g^{\mathrm{phase}}_k\,g^{\mathrm{conflict}}_k\,g^{\mathrm{face}}_k\,
 g^{\mathrm{subtle}}_k\,u_k,\qquad g_k\in[0,1],
 \label{eq:gate}
\end{equation}
with neutral factors for channels to which a condition does not apply. Phase denotes the
training-step warmup, face gating controls face-dependent evidence, conflict gating
attenuates disagreement between semantic providers, and subtlety gating restrains
intensity-sensitive cues. Motion naturalness uses its observation confidence.
These controls are computed, not learned (Appendix~\xdocref{app:gates}).

\paragraph{Evidence-dependent trust.}
The phase factor schedules reward evidence according to training progress, separately from
the within-video expression window. Face gating limits channels requiring facial
evidence; conflict and subtlety gating temper disagreeing semantic assessments and
intensity-sensitive cues. Their provider-specific application changes relative
influence; a shared positive attenuation would largely cancel under normalization.

\paragraph{Composition with experience.}

For a fixed base weight $w_k^0$, define the effective contribution
$a_k(x,p)=w_k^0 g_k(x,p)m_k(r(x,p))$. Reflective Reward Gating uses $m_k=1$;
EmoReward uses the fixed table produced by Evolve. With
$\mathcal A=\{k:a_k>\epsilon_g\}$, active-provider normalization gives
\begin{equation}
 R_{\mathrm{emo}}(x,p)=
 \frac{\sum_{k\in\mathcal A} a_k(x,p)s_k(x,p)}{\sum_{k\in\mathcal A}a_k(x,p)+\epsilon}.
 \label{eq:active_fusion}
\end{equation}
Invalid evidence receives no mass. A common positive factor cancels except through the
stabilizer; calibration acts through relative provider contributions and differences
between videos. Regime $r$ selects a table entry; gate $g_k$ controls current evidence validity.
A provider favored by memory still contributes nothing when its current evidence
is invalid. With no memory entry, unit multipliers recover Reflect-only fusion;
the scoring interfaces and the policy objective remain the same.

\begin{algorithm}[t]
\caption{RewardMemory construction from reflective reward traces}
\label{alg:jit}
\begin{algorithmic}[1]
\Require Traces $\mathcal M$ from a completed reference run, provider count $K$, temperature $\tau$, log-multiplier cap $c$, minimum support $n_{\min}$
\For{each context and the global fallback}
\State Select matching traces; use the global set if the context has fewer than $n_{\min}$ traces
\For{each provider $k$}
\State $\rho_k\gets\operatorname{Corr}(s_{i,k},y_i)$; use zero for constant scores
\EndFor
\State $\bar\rho\gets K^{-1}\sum_j\rho_j$
\For{each provider $k$}
\State $m_k\gets\exp(\operatorname{clip}((\rho_k-\bar\rho)/\tau,-c,c))$
\EndFor
\State Store the context multipliers
\EndFor
\State Export once; freeze the table for a subsequent generator-training run
\end{algorithmic}
\end{algorithm}

\subsection{Evolve: RewardMemory}
\label{sec:evolve}

RewardMemory treats reward allocation as a decision over emotional context $z$: the
providers are actions, their fusion weights form a policy, and a stored outcome $y$
supplies a non-parametric reliability signal. Experience is collected during a
\emph{completed reference training run} with Reflective Reward Gating. Each trace
$m_i=(z_i,\{s_i^k\},\{g_i^k\},w_i,R_i,y_i)$ records context, scores, gates, weights,
and outcomes. Evolve groups these traces into a regime neighborhood $\mathcal N(z)$,
with a global fallback for insufficient support, and computes
\begin{equation}
\begin{aligned}
 \rho_k(z)&=\frac{\operatorname{Cov}_{i\in\mathcal N(z)}(s_i^k,y_i)}
 {\sigma_{s^k}\sigma_y+\epsilon},\\
 A_k^{\mathrm{rel}}(z)&=\rho_k(z)-\tfrac{1}{K}\sum_j\rho_j(z).
\end{aligned}
 \label{eq:rel_adv}
\end{equation}
Reliability uses raw provider scores; constant scores receive zero correlation.
It measures agreement with the reference outcome, not independent emotional fidelity.
The centered advantage tilts the base allocation through bounded multipliers:
\begin{equation}
\begin{aligned}
 \Delta_k&=\operatorname{clip}(A_k^{\mathrm{rel}}/\tau,-c,c),\\
 w_k^{\mathrm{RM}}(z)&=\frac{w_k^0e^{\Delta_k}}{\sum_jw_j^0e^{\Delta_j}}.
\end{aligned}
 \label{eq:jit_weight}
\end{equation}
This is the closed-form optimum of a KL-regularized allocation with clipped utilities
(Appendix~\xdocref{app:derivation}). The exported $m_k(r)=e^{\Delta_k}$ compose with current
instance gates in Eq.~\ref{eq:active_fusion}. The table is fixed throughout the next
training run: gradients update the video generation model alone, and the table is read-only. Algorithms~\ref{alg:emoreward}
and~\ref{alg:jit} separate reward computation from between-run memory construction.
Appendix~\xdocref{app:retrieval} explains context grouping and table construction;
Appendix~\xdocref{app:walkthrough} provides a trace-based worked example.

%% file: sections/04_experiment.tex
\section{Experiments}
\label{sec:experiments}

\begin{figure}[t]
\centering
\includegraphics[width=.85\textwidth]{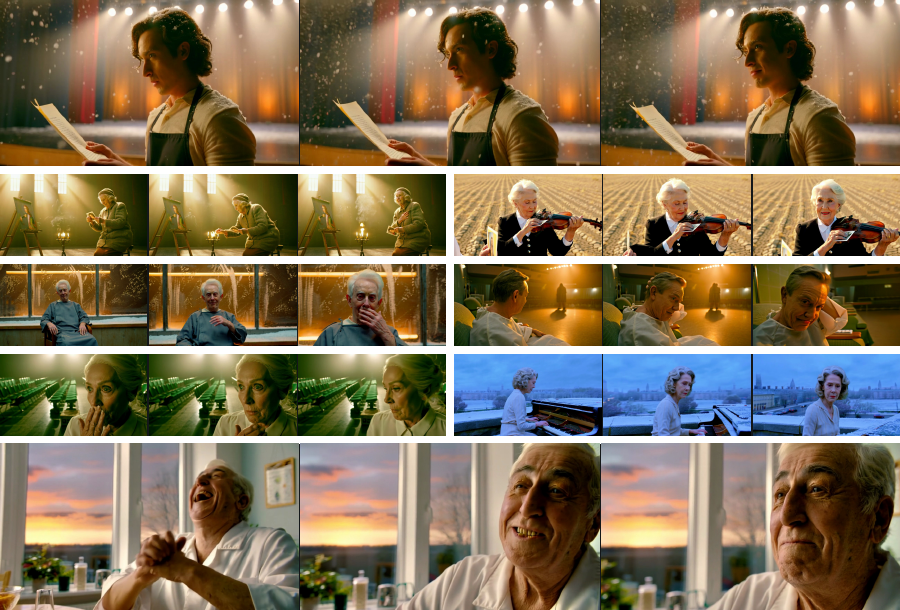}
\caption{\textbf{Emotional expression in generated videos.}
Eight clips are shown at three moments each. \textbf{(Top)} A performer looks up
from a letter. \textbf{(Middle rows)} Gaze, gesture, and expression convey emotion
in scenes such as a violin performance and a rooftop piano scene.
\textbf{(Bottom)} Laughter fades into a quiet gaze.}
\label{fig:video_showcase}
\end{figure}

\begin{figure}[!t]
\centering
\input{sections/tables/agentic_comparison}
\par\medskip
\input{sections/rl_qualitative}
\end{figure}

\input{sections/tables/video_main}

\paragraph{Experimental settings.}
We use HunyuanVideo~1.5 with DanceGRPO, FlowGRPO, and SAGE-GRPO, fusion
$\alpha=\beta=0.5$, batch size $2$, and group size $G=4$.
We use Qwen3-VL to create emotional prompts for training and use Claude to newly create 302 prompts in same style for evaluation (Appendix~\xdocref{app:prompt_construction}).
GT denotes the pretrained video generation model without RL.
The controlled seven-policy comparison retains its separate evaluation protocol.
VideoAlign supplies $R_{\mathrm{vid}}$ and the reference outcome $y$.
Memory calibration uses $K=7$, $\tau=0.5$, a log-multiplier cap $c=1$, and minimum context support $n_{\min}=20$.

\subsection{Main Experiments}
\label{sec:exp1}

\paragraph{Quantitative evaluation.}
We compute Directional Emotion Accuracy (DEA) and Target Emotion Rank Score
(TERS) with a frozen CLIP ViT-H/14 encoder~\citep{schuhmann2022laionb}, which is different from the separate CLIP-AUTT
model~\citep{zeeshan2026clipautt} used as the facial reward during training (details in Appendix~\xdocref{app:dea}).
Table~\ref{tab:exp1} compares EmoReward with each RL algorithm's base reward
and includes the pretrained video generation model (GT) as a reference.
All three RL baselines score below GT on DEA and TERS.
EmoReward improves both metrics under every RL algorithm, with TERS gains
of $0.016$--$0.031$. FlowGRPO and SAGE-GRPO also surpass GT on both metrics.
The gains have the same direction with a second frozen CLIP encoder.
FlowGRPO and SAGE-GRPO improve on all four
VideoAlign measures. 
% DanceGRPO shows small gains in visual and motion quality alongside declines in text alignment and Overall. 
Both visual metrics increase with EmoReward under SAGE-GRPO, with mixed
changes across the other algorithms. Figure~\ref{fig:rl_qualitative} illustrates
the changes in emotional expression across all three algorithms.

\paragraph{Qualitative analysis.}
\label{sec:qualitative_analysis}

Figure~\ref{fig:rl_qualitative} compares base and EmoReward results for each RL algorithm with matched prompts and seeds.
For DanceGRPO, the man stays bowed with his face covered with better visual effects, and for FlowGRPO the reporter's strained expression is clearer from a more frontal angle, while under SAGE-GRPO the musician's smile fades into a fixed stare. These examples complement the clip-level emotion scores by showing the order in which the visible cues emerge.

\subsection{Ablations and Mechanism Analysis}
\label{sec:exp2}
\label{sec:exp3}
\paragraph{Emotion-category analysis.}
Figure~\ref{fig:emotion_categories} compares GT with SAGE-GRPO plus EmoReward across eight emotion categories. 
All four VideoAlign measures increase in every category, while PickScore and CLIPScore vary. 
This comparison captures the effect of post-training as a whole.

\paragraph{Reward-policy ablation.}
We compare seven reward policies with a fixed backbone and training budget
(Table~\ref{tab:agentic}). RewardMemory leads in Sel. ($0.428$), GD, and GR,
while Reflect leads in EF, the ImageReward preference proxy. SNR$^*$ measures
relative dispersion. The predefined core and emo scores are descriptive
summaries, since min--max scaling can magnify small differences
(Appendix~\xdocref{app:gating_metrics}).

\paragraph{Reward-provider pool.}
Figure~\ref{fig:provider_pool} compares VideoAlign, semantic emotion, motion
naturalness, and a balanced pool. Weights and curricula co-vary with the pool
(Appendix~\xdocref{app:ablation}). Seven channels enter memory calibration.
Chromatic and kinetic cues supply auxiliary readouts.

\paragraph{Memory-conditioned trust.}
The largest global multipliers go to macro ($1.833$) and motion ($1.510$).
Three constant channels receive $0.778$. The gates still exclude invalid
evidence. Macro shares an evaluator with the reference outcome, so its
correlation reflects objective agreement. Appendices~\xdocref{app:retrieval} and~\xdocref{app:walkthrough} describe and illustrate the allocation.

\subsection{Generalization to image backbones}
\label{sec:exp4}

We test SD3.5 and FLUX across three RL algorithms with EmoReward. 
EmoReward improves HPSv3 for SD3.5, while FLUX's baseline still leads on the same metric. We tested on serveal prompts and find EmoReward enhance the text following ability and reduce deformities.
Appendix~\xdocref{app:robust} details the settings and qualitative comparisons.

\subsection{Human evaluation}
\label{sec:human}

\begin{figure}[H]
\centering
\begin{minipage}[t]{.49\linewidth}
\vspace{0pt}\centering
\includegraphics[width=\linewidth]{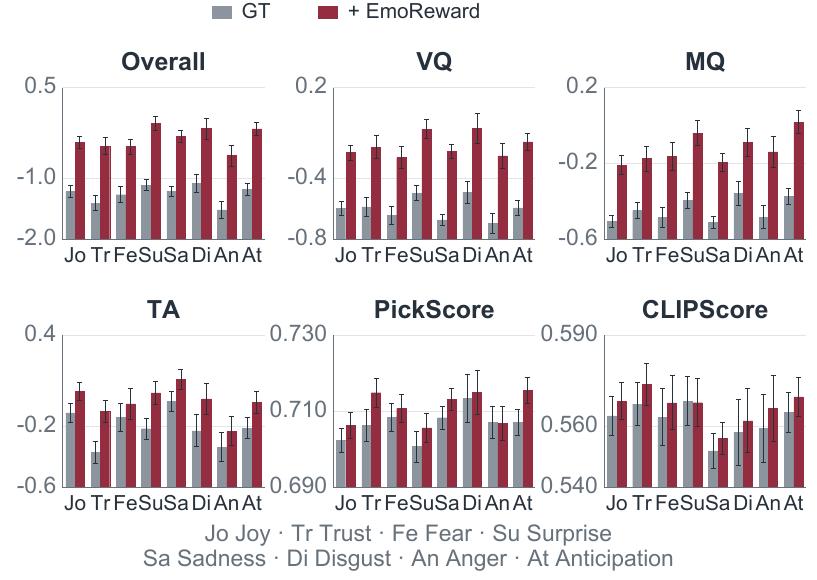}
\caption{\textbf{Emotion-category quality.} GT (gray) versus SAGE-GRPO with EmoReward (red); mean $\pm$ SEM. Axes are truncated; categories overlap within the evaluated set.}
\label{fig:emotion_categories}
\end{minipage}\hfill
\begin{minipage}[t]{.49\linewidth}
\vspace{0pt}\centering
\includegraphics[width=\linewidth]{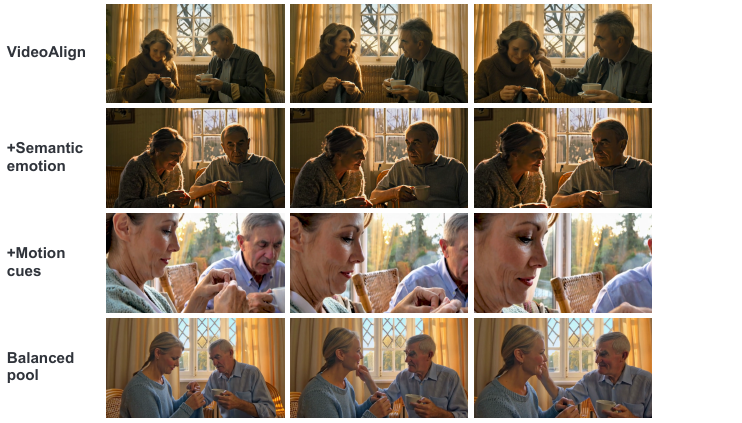}
\caption{\textbf{Ablation of provider-pool construction.} Three frames per clip compare four configurations on the same prompt. See Appendix~\xdocref{app:ablation} for quantitative results.}
\label{fig:provider_pool}
\end{minipage}
\par\vspace{14pt}
\begin{minipage}{\textwidth}
\centering\small
\setlength{\tabcolsep}{3pt}
\captionof{table}{\textbf{Human evaluation.} Rating percentages from 18 participants on 30 prompts.
Each method and criterion block contains 540 judgments. Bold marks the highest Good share
per criterion; percentages are rounded to one decimal place.}
\label{tab:human}
\resizebox{\linewidth}{!}{\input{sections/tables/human_results}}
\end{minipage}
\end{figure}

Eighteen participants rated three methods on 30 prompts for emotion fidelity, temporal coherence, and visual quality. 
They used the same Bad/Normal/Good scale for each criterion. SAGE-GRPO receives more Good and more Bad ratings than GT on all three criteria. 
EmoReward has the highest Good shares
($60.0\%$, $54.3\%$, $57.2\%$), compared with $53.0\%$, $51.5\%$, and $45.0\%$ for SAGE-GRPO, and the lowest Bad shares (Table~\ref{tab:human}).
The same participants rated multiple prompts across methods, and Appendix~\xdocref{app:human} details the protocol.

%% file: sections/tables/agentic_comparison.tex
\begin{minipage}{\textwidth}
\centering\small
\setlength{\tabcolsep}{3pt}
\captionof{table}{\textbf{Reward-policy ablation under SAGE-GRPO.} Controlled policies share the backbone and training budget. Citations identify components. SNR$^*$ is relative dispersion. Bold marks diagnostic and pillar maxima. Core and emo are shown for reference.}
\label{tab:agentic}
\begin{tabular*}{\linewidth}{@{\extracolsep{\fill}}lrrrrrrr@{}}\toprule
Reward method & SNR$^*$ & Sel.$\uparrow$ & EF$\uparrow$ & GD$\uparrow$ & GR$\uparrow$ & core$\uparrow$ & emo$\uparrow$ \\\midrule
VideoAlign \citep{liu2025videoalign} & 0.378 & 0.307 & 0.518 & 0.000 & 0.197 & 0.238 & 0.308 \\
Single provider \citep{zhao2025r1omni} & 2.403 & 0.314 & 0.454 & 0.305 & 0.147 & 0.302 & 0.340 \\
Random routing \citep{wu2023autogen} & 1.105 & 0.340 & 0.000 & 0.109 & 0.146 & 0.085 & 0.064 \\
Round-robin \citep{wu2023autogen} & 1.217 & 0.376 & 0.291 & 0.126 & 0.381 & 0.266 & 0.272 \\
Fixed fusion \citep{xu2024visionreward} & 0.773 & 0.371 & 0.026 & 0.060 & 0.653 & 0.246 & 0.191 \\
\textbf{EmoReward (w/o RewardMemory)} & 0.451 & 0.293 & \textbf{1.000} & 0.011 & 0.000 & 0.337 & 0.503 \\
\textbf{EmoReward (w/ RewardMemory)} & \textbf{7.022} & \textbf{0.428} & 0.767 & \textbf{1.000} & \textbf{1.000} & \textbf{0.922} & \textbf{0.884} \\
\bottomrule\end{tabular*}
\end{minipage}

%% file: sections/rl_qualitative.tex
\begin{minipage}{\textwidth}
\centering
\includegraphics[width=.88\linewidth]{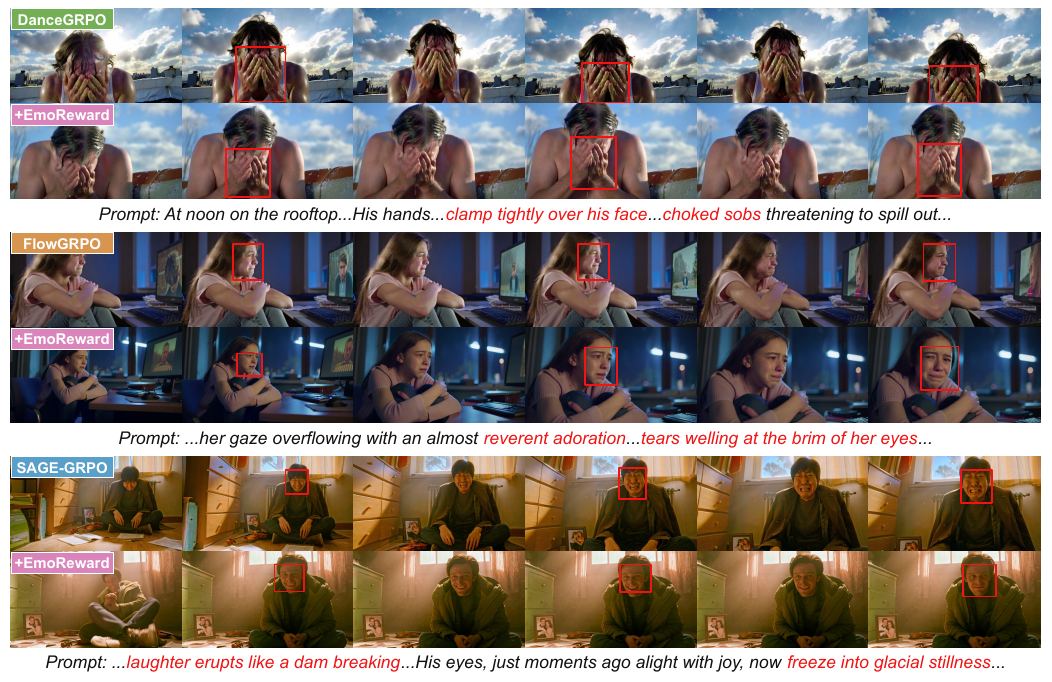}
\captionof{figure}{\textbf{Qualitative comparison across video RL algorithms.}
Each pair shares a prompt and seed. Red boxes mark the hands and facial cues.
\textbf{(Top)} A bowed head and hands held over the face convey grief.
\textbf{(Middle)} The reporter's expression tightens as she reacts to the screen.
\textbf{(Bottom)} A broad smile gives way to a fixed stare.}
\label{fig:rl_qualitative}
\end{minipage}

%% file: sections/tables/video_main.tex
% Final delivery table - 0916. Requires \usepackage[table]{xcolor}, booktabs.
% Row shading: GT = light grey, +EmoReward = light blue. Values: mean $\pm$ SEM.
% Each guided row uses full EmoReward. The DanceGRPO row is recomputed on the same 302 prompts.
\definecolor{gtgrey}{gray}{0.90}
\definecolor{emoblue}{RGB}{247,234,237}
\begin{table}[t]\centering\small
\setlength{\tabcolsep}{3pt}
\caption{\textbf{EmoReward across video RL algorithms.}
DEA uses the 261 prompts with a usable opposite emotion. TERS ranks the target
emotion on all 302 prompts. All metrics report mean $\pm$ SEM across prompts. GT is the pretrained no-RL reference. Bold marks an
increase over its corresponding RL baseline.}
\label{tab:exp1}
\resizebox{\linewidth}{!}{%
\begin{tabular}{@{}lcccccccc@{}}
\toprule
& \multicolumn{2}{c}{Emotion Metrics} & \multicolumn{4}{c}{VideoAlign} & \multicolumn{2}{c}{Visual Metrics} \\
\cmidrule(lr){2-3}\cmidrule(lr){4-7}\cmidrule(lr){8-9}
Method & DEA & TERS & Overall & VQ & MQ & TA & PickScore & CLIPScore \\
\midrule
\rowcolor{gtgrey} GT (reference) & 0.655{\scriptsize$\pm$0.029} & 0.7904{\scriptsize$\pm$0.0145} & -1.1938{\scriptsize$\pm$0.0520} & -0.5899{\scriptsize$\pm$0.0255} & -0.4289{\scriptsize$\pm$0.0220} & -0.1750{\scriptsize$\pm$0.0363} & 0.7059{\scriptsize$\pm$0.0018} & 0.5623{\scriptsize$\pm$0.0037} \\
\midrule
DanceGRPO & 0.605{\scriptsize$\pm$0.030} & 0.7640{\scriptsize$\pm$0.0153} & -1.2812{\scriptsize$\pm$0.0513} & -0.6392{\scriptsize$\pm$0.0230} & -0.3867{\scriptsize$\pm$0.0235} & -0.2554{\scriptsize$\pm$0.0357} & 0.7177{\scriptsize$\pm$0.0018} & 0.5675{\scriptsize$\pm$0.0036} \\
\rowcolor{emoblue} \quad + EmoReward & \textbf{0.632}{\scriptsize$\pm$0.030} & \textbf{0.7824}{\scriptsize$\pm$0.0149} & -1.3930{\scriptsize$\pm$0.0559} & \textbf{-0.6279}{\scriptsize$\pm$0.0284} & \textbf{-0.3833}{\scriptsize$\pm$0.0243} & -0.3818{\scriptsize$\pm$0.0340} & 0.6986{\scriptsize$\pm$0.0017} & 0.5599{\scriptsize$\pm$0.0037} \\
\midrule
FlowGRPO & 0.621{\scriptsize$\pm$0.030} & 0.7862{\scriptsize$\pm$0.0144} & -0.7719{\scriptsize$\pm$0.0569} & -0.3567{\scriptsize$\pm$0.0290} & -0.2753{\scriptsize$\pm$0.0271} & -0.1398{\scriptsize$\pm$0.0361} & 0.7061{\scriptsize$\pm$0.0019} & 0.5634{\scriptsize$\pm$0.0036} \\
\rowcolor{emoblue} \quad + EmoReward & \textbf{0.670}{\scriptsize$\pm$0.029} & \textbf{0.8018}{\scriptsize$\pm$0.0142} & \textbf{-0.5376}{\scriptsize$\pm$0.0556} & \textbf{-0.3398}{\scriptsize$\pm$0.0291} & \textbf{-0.2302}{\scriptsize$\pm$0.0275} & \textbf{0.0323}{\scriptsize$\pm$0.0356} & \textbf{0.7115}{\scriptsize$\pm$0.0018} & 0.5605{\scriptsize$\pm$0.0037} \\
\midrule
SAGE-GRPO & 0.617{\scriptsize$\pm$0.030} & 0.7881{\scriptsize$\pm$0.0143} & -0.8269{\scriptsize$\pm$0.0532} & -0.4106{\scriptsize$\pm$0.0261} & -0.3275{\scriptsize$\pm$0.0263} & -0.0888{\scriptsize$\pm$0.0360} & 0.7091{\scriptsize$\pm$0.0019} & 0.5624{\scriptsize$\pm$0.0036} \\
\rowcolor{emoblue} \quad + EmoReward & \textbf{0.674}{\scriptsize$\pm$0.029} & \textbf{0.8193}{\scriptsize$\pm$0.0132} & \textbf{-0.2624}{\scriptsize$\pm$0.0600} & \textbf{-0.1710}{\scriptsize$\pm$0.0304} & \textbf{-0.0926}{\scriptsize$\pm$0.0324} & \textbf{0.0012}{\scriptsize$\pm$0.0364} & \textbf{0.7107}{\scriptsize$\pm$0.0018} & \textbf{0.5663}{\scriptsize$\pm$0.0035} \\
\bottomrule
\end{tabular}}
\end{table}

%% file: sections/tables/human_results.tex
\begin{tabular}{lrrrrrrrrr}
\toprule
& \multicolumn{3}{c}{Emotion Fidelity} & \multicolumn{3}{c}{Temporal Coherence} & \multicolumn{3}{c}{Visual Quality} \\
\cmidrule(lr){2-4}\cmidrule(lr){5-7}\cmidrule(lr){8-10}
Method & Bad $\downarrow$ & Normal & Good $\uparrow$ & Bad $\downarrow$ & Normal & Good $\uparrow$ & Bad $\downarrow$ & Normal & Good $\uparrow$ \\
\midrule
GT & 21.9\% & 34.8\% & 43.3\% & 23.5\% & 37.8\% & 38.7\% & 25.2\% & 38.5\% & 36.3\% \\
SAGE-GRPO & 31.5\% & 15.6\% & 53.0\% & 28.9\% & 19.6\% & 51.5\% & 30.9\% & 24.1\% & 45.0\% \\
SAGE-GRPO + EmoReward & \textbf{15.4}\% & 24.6\% & \textbf{60.0\%} & \textbf{20.2}\% & 25.6\% & \textbf{54.3\%} & \textbf{15.9}\% & 26.9\% & \textbf{57.2\%} \\
\bottomrule
\end{tabular}

%% file: sections/05_conclusion.tex
\section{Conclusion}
In this paper, we proposed EmoReward, a self-evolving reward calibration framework
for emotional video generation reinforcement learning. 
EmoReward combines instance-level Reflective Reward Gating with RewardMemory, which transfers provider-reliability estimates from a completed training run to subsequent optimization without updating reward-model parameters. Experiments across video optimizers, complemented by ablations,
image-transfer analyses, and human ratings, support the utility of adaptive calibration
while identifying objective-dependent trade-offs. These findings suggest that
completed-run experience can inform the allocation of trust among heterogeneous
reward providers, providing a foundation for further research on reward-system evolution across training.

%% file: sections/06_llm_usage.tex
\section*{LLM Usage Statement}
OpenAI ChatGPT and Codex assisted with manuscript editing and organization, checks of
the mathematical exposition, and preparation of figure layouts and LaTeX source.
We used Qwen3-VL to create emotional prompts for training and Claude to author
a separate dataset of prompts for evaluation. The main comparison uses 302
prompts from this dataset (Appendix~\xdocref{app:prompt_construction}).
The vision--language and generative models used in the reward pipeline and experiments
are described in the paper and supplement. Experimental comparison images and video
frames are outputs of the evaluated models; no additional media were generated to
illustrate experimental outcomes. The authors are responsible for the scientific
claims, evidence, and final manuscript.

%% file: sections/10_appendix.tex
\input{sections/appendix_A_limits}
\input{sections/appendix_B_prompt_construction}
\input{sections/11_human_protocol}
\input{sections/appendix_B_providers}

\input{sections/appendix_C_gates}

\input{sections/appendix_D_memory}
\input{sections/appendix_E_derivation}
\input{sections/appendix_F_metrics}

\input{sections/appendix_G_ablations}

\input{sections/appendix_H_walkthrough}
\input{sections/appendix_I_images}

\input{sections/appendix_K_videos}

%% file: sections/appendix_A_limits.tex
\section{Limitations and Future Work}
\label{app:limitations}
\paragraph{Beyond emotional video generation.}
EmoReward studies a training-adaptive reward system that transfers evidence about
provider reliability between generator-training runs without updating reward-model
parameters. The present evaluation focuses on emotion, where different scenes call
for different combinations of facial, gestural, semantic, and motion evidence.
A broader application would require a task-appropriate provider pool and a meaningful
outcome for calibration. Action fidelity and audio--visual correspondence are natural
visual extensions; language models and tool-use agents could similarly combine
semantic judges, execution checks, and trajectory-level assessments. Whether the same
calibration rule transfers effectively remains an empirical question, especially when
providers share errors or outcomes favor only part of the task.

\paragraph{Iterative evolution across training runs.}
A completed training run can be treated as a unit of experience for its successor.
Repeating this procedure could produce a sequence of generators and reward-allocation
tables, with each generation contributing new traces. Our experiments evaluate a
reference-to-calibrated transition, not sustained improvement across many generations.
Such an extension must address distribution shift, stale evidence, and feedback loops
that reinforce a proxy's errors. Retaining diverse historical traces and evaluating
held-out outcomes would be important tests of whether repeated calibration remains useful.

\paragraph{Model-level human feedback.}
The trace representation also suggests a way to carry human assessments of a completed
model into subsequent training. For example, a reviewer who finds movement too weak
could request greater emphasis on suitable motion providers in the affected contexts;
excessive motion could motivate the opposite adjustment. This would complement
sample-level preference feedback with structured assessments of a model's final
behavior, translated into bounded changes to the next run's reward allocation.
Mapping subjective feedback to providers and regimes, reconciling different raters,
and preventing compensating failures remain open design questions. The ordinal study
reported here evaluates outputs; it does not implement this proposed feedback mechanism.

%% file: sections/appendix_B_prompt_construction.tex
\section{Training and Evaluation Prompts Preparation}
\label{app:prompt_construction}

\paragraph{Training corpus.}
We use Qwen3-VL-30B-A3B~\citep{qwen3technicalreport,bai2025qwen3vl} to write
emotional video prompts. 
Our 62-category taxonomy draws on Plutchik's eight
primary emotions and their derivatives, Cowen and Keltner's 27--28-category
semantic-space studies, Shaver's three-level hierarchy as presented in
Parrott's collection, Ekman and Friesen's Facial Action Coding System (FACS),
Hume AI's facial expression model, and GoEmotions' 28 labels including
neutral~\citep{plutchik2001emotions,cowen2017selfreport,cowen2020face,
shaver1987emotion,parrott2001emotions,ekman1978facs,demszky2020goemotions}.
Each category has a Chinese and an English label and specifies valence, arousal, facial action units, body cues, and temporal dynamics. 
We assign a seed to each row, sample a category from that seed, and use the corresponding label to condition prompt generation. 
The resulting corpus contains 10,000 distinct prompts to diversify rollouts across training. 
Each prompt is used at most once in a complete online on-policy training run.

\paragraph{Example.}
For \mbox{row\_seed = 478163327}, we draw \textbf{Love/Affection}, with positive valence and medium arousal. Its required facial
actions are AU6 (cheek raiser) and AU12 (lip corner puller). The body cues
are intimate distance (0--45~cm), gentle stroking, and a soft, nonthreatening
gaze. The target expression develops slowly and sustains a sense of warmth.
The system prompt below is shown in English translation.

\begin{quote}\small
Write a visual description for text-to-video generation in Simplified Chinese.
Output a single paragraph with no title, list, or quotation marks. Center the
scene on people, with visible facial expressions, gaze, lip movements, and
body language. Show subtle shifts in emotion and microexpressions. Convey
love or affection through the scene and the performances and avoid repeating
the emotion label. Write roughly 180--350 Chinese characters. Use cinematic
language to describe one or two of framing, camera angle, lighting, and camera
movement, and establish the atmosphere of the setting. Exclude prohibited
content and excessively graphic violence. Output only the Chinese paragraph,
with no explanatory preface or closing note.

\smallskip
Random seed for inspiration only (do not reproduce the number)\quad 156744132
\end{quote}

Showing the emotion through acting gives the facial and body reward providers
visible cues to score. Varying the inspiration seed encourages different scenes
when the same category is sampled again. An example rollout prompt is shown
below in English.

\begin{quote}\small
The slanting sunlight of dusk filters through the old window lattice, bathing
the room in a warm orange glow. A middle-aged couple sits on worn wicker
chairs; she is bent over, mending an old sweater, her fingers deftly guiding
the needle and thread. He watches her quietly, his fingers absently tracing the
rim of his cup\ldots{} Suddenly, she looks up at him, a faint smile lifting the
corners of her lips, fine lines crinkling at the edges of her eyes---gentle
ripples of time-worn tenderness\ldots{} he reaches out, gently brushing a
strand of her hair, his touch slow and reverent\ldots{} they share a glance,
a silent understanding flowing between them. The camera
slowly pushes in, focusing on the delicate smile lines at the corner of her eye
and the quiet, shimmering light deep within his gaze.
\end{quote}

\paragraph{Evaluation corpus.}
We used Claude to create a separate emotional prompt dataset for evaluation,
following the training corpus's emotion specifications and cinematic style.
These prompts were held out from RL training. Table~\xdocref{tab:exp1} uses the
same 302-prompt subset for each RL algorithm and its matched base and EmoReward
runs.
Appendix~\ref{app:dea} describes how the automatic
metrics are aggregated, including DEA's requirement for a usable opposite emotion.

%% file: sections/11_human_protocol.tex
\section{Human Evaluation: Ordinal Ratings}
\label{app:human}

\paragraph{Study and comparison groups.}
The study includes 18 participants and 30 prompts, with one video per prompt from
each of three methods: the pretrained video generation model without RL (GT), SAGE-GRPO, and
SAGE-GRPO with EmoReward. GT denotes the pretrained model baseline.
Participants consented to take part in the study.
Participants assessed each output separately on three criteria using Bad, Normal,
or Good. The supplied aggregate results contain 540 judgments for each method--criterion
combination, consistent with the full participant--prompt product. Across all three
methods and criteria, there are 4,860 categorical judgments. These ordinal counts
are separate from the earlier pairwise preference protocol; pairwise choices are
not converted into categorical ratings.

\paragraph{Criteria.}
\emph{Emotion fidelity} concerns how convincingly the video conveys the prompt's intended
affect, including whether expression and gesture fit its emotional tone.
\emph{Temporal coherence} concerns continuity of expression, identity, and movement
through the clip. \emph{Visual quality} concerns perceptual detail, anatomy, and visible
artifacts. The categories distinguish poor, intermediate, and good quality on each
axis. They are rated separately rather than collapsed into a single preference or
an assumption that stronger emotional intensity is always better.

% \begin{figure}[!ht]
% \centering
% \includegraphics[width=\linewidth]{figures/user_study/ordinal_ui.pdf}
% \par\vspace{4pt}\noindent\parbox{\linewidth}{\small
% \textbf{Interface for ordinal human evaluation.} The interface presents the prompt and generated video alongside separate ratings of emotion fidelity, temporal coherence, and visual quality. Each criterion uses the same three ordered categories (Bad, Normal, Good), allowing the dimensions to be assessed independently. This reconstruction illustrates the reported rating protocol; it is not a record of the original data-collection interface.}
% \end{figure}
\begin{figure}[!ht]
  \centering
  \includegraphics[width=\linewidth]{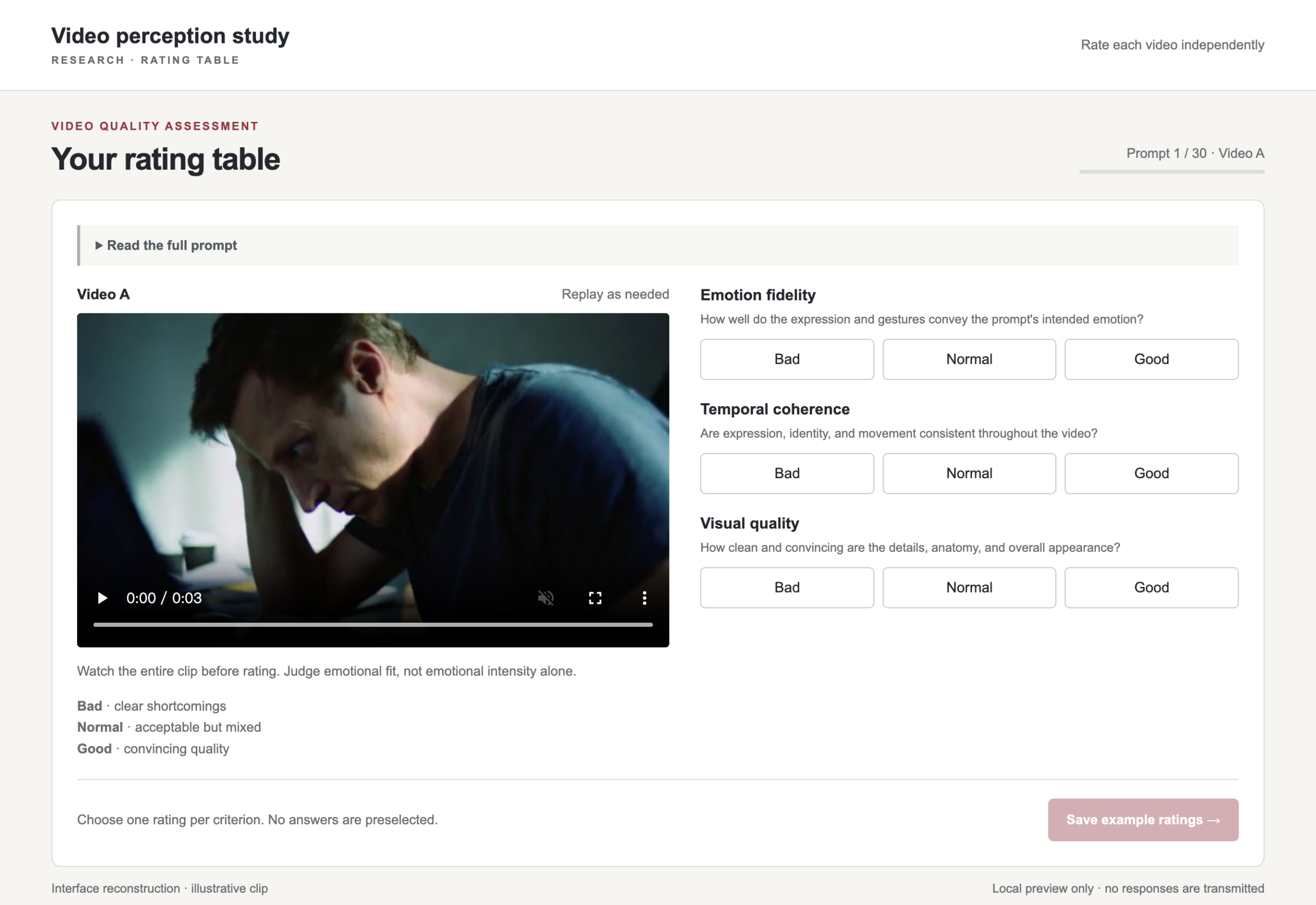}
  \caption{\textbf{Interface for ordinal human evaluation.}
  The interface presents the prompt and generated video with independent ratings of emotional fidelity, temporal coherence, and visual quality, each using three ordered categories: Bad, Normal, and Good. This reconstruction illustrates the reported rating protocol rather than the original data-collection interface.}
  \label{fig:ordinal_ui}
\end{figure}

\paragraph{Aggregation and rounding.}
For method $m$, criterion $c$, and rating category $b$, the percentage is:
\begin{equation}
 \widehat p_{m,c,b}=100\,\frac{N_{m,c,b}}{\sum_{b'\in\{\mathrm{Bad,Normal,Good}\}}N_{m,c,b'}},
 \qquad \sum_{b'}N_{m,c,b'}=540.
\end{equation}
Table~\xdocref{tab:human} rounds each percentage independently to one decimal place and
bolds only the highest Good percentage within each criterion. Rounding may make a
three-category total differ from $100.0\%$ by $0.1$ percentage points. The raw integer
counts are given in Table~\ref{tab:human_counts} and were not rounded before normalization. All nine aggregate blocks have the stated denominator; the totals alone
do not establish participant-level completeness or the absence of duplicate records.

\input{sections/tables/human_counts}

\paragraph{Reading the distributions.}
Relative to SAGE-GRPO, EmoReward's Good share increases by $7.0$, $2.8$, and $12.2$
percentage points for emotion fidelity, temporal coherence, and visual quality,
respectively. These differences are computed from the unrounded counts.
Its Bad share falls from $31.5\%$ to $15.4\%$ for emotion fidelity, from $28.9\%$
to $20.2\%$ for temporal coherence, and from $30.9\%$ to $15.9\%$ for visual quality.
The complete distribution matters: compared with GT, SAGE-GRPO has more Good ratings
but also more Bad ratings on every criterion, so a Good-only summary would omit
part of the observed pattern.

\paragraph{Statistical scope.}
The present report uses the supplied aggregated counts. It does not treat repeated
judgments from the same participant or prompt as independent observations. Participant-
and prompt-level records are needed for uncertainty estimates that respect both sources
of dependence; no confidence interval or significance claim is inferred from the
aggregate margins. The aggregates do not document recruitment, presentation order,
blinding, exclusions, or prompt-selection provenance. Those collection details must
be established from the study records rather than inferred from the rating totals.

%% file: sections/tables/human_counts.tex
\begin{table}[ht]
\centering\small
\caption{\textbf{Aggregate ordinal counts.} Each method--criterion block contains 540 ratings. The counts reproduce the percentages in main-paper Table~\xdocref{tab:human}.}
\label{tab:human_counts}
\begin{tabular*}{\linewidth}{@{\extracolsep{\fill}}lrrrrrrrrr@{}}
\toprule
& \multicolumn{3}{c}{Emotion fidelity} & \multicolumn{3}{c}{Temporal coherence} & \multicolumn{3}{c}{Visual quality} \\
\cmidrule(lr){2-4}\cmidrule(lr){5-7}\cmidrule(lr){8-10}
Method & Bad & Normal & Good & Bad & Normal & Good & Bad & Normal & Good \\
\midrule
GT & 118 & 188 & 234 & 127 & 204 & 209 & 136 & 208 & 196 \\
SAGE-GRPO & 170 & 84 & 286 & 156 & 106 & 278 & 167 & 130 & 243 \\
SAGE-GRPO + EmoReward & 83 & 133 & 324 & 109 & 138 & 293 & 86 & 145 & 309 \\
\bottomrule
\end{tabular*}
\end{table}

%% file: sections/appendix_B_providers.tex
\section{Act: Provider Interface and Reference Scoring Rules}
\label{app:primitives}
This appendix specifies the scoring operators used by the provider interface in
Eq.~\xdocref{eq:primitive_score}. The six primitive \emph{families} (facial
expression; semantic emotion understanding; body and temporal evidence; chromatic atmosphere;
kinetic intensity; motion naturalness) are instantiated by nine observation channels;
Table~\ref{tab:provider_map} maps each family to its provider(s), the backing estimator, and the
defining equation. Throughout, $T$ is the number of sampled frames,
$\mathcal{F}\subseteq\{1,\dots,T\}$ the face-detected subset, $\omega_t$ the EmoScript time
weight (per-window kernels
$\psi_\ell(t)=\mathbf{1}[t\in\omega_\ell]/(\sum_{t'}\mathbf{1}[t'\in\omega_\ell]+\epsilon)$ over
$\Omega=\{[t_s^\ell,t_e^\ell]\}$), and $\kappa(x;\sigma)=\exp(-\max(0,x)/\sigma)$ a bounded
exponential score. Each provider $k$ returns a score $s_k\in[0,1]$ and a validity signal $u_k$.

\begin{table}[t]
\centering
\small
\caption{\textbf{Primitive family, provider, backing estimator, and defining equation.} The two
appearance channels (chromatic, kinetic) are always-on lightweight signals; the remaining seven
form the memory support whose per-provider reliability is reported in the calibration analysis
(Table~\ref{tab:exp3_jit}).}
\label{tab:provider_map}
\begin{tabular}{@{}llll@{}}
\toprule
Family & Provider & Estimator / signal & Eq. \\
\midrule
Facial & AU matching & Facial action-unit head & \ref{eq:app_micro} \\
Facial & CLIP-AUTT & Temporal AU--text similarity & \ref{eq:app_clip} \\
Semantic & R1-Omni & Valence--arousal estimation & \ref{eq:app_r1} \\
Semantic & Macro & Alignment and AU compliance & \ref{eq:app_macro} \\
Body & Body cues & Body-marker satisfaction & \ref{eq:app_body} \\
Temporal & Peak alignment & AU peak versus target window & \ref{eq:app_temporal} \\
Appearance & Chromatic & Saturation and warm hue & \ref{eq:app_chromatic} \\
Appearance & Kinetic & Inter-frame differences & \ref{eq:app_kinetic} \\
Motion & Naturalness & Optical-flow consistency & \ref{eq:app_motion} \\
\bottomrule
\end{tabular}
\end{table}

\subsection{Facial expression: AU matching and expression similarity}
\label{app:prim_face}

\paragraph{AU matching (\emph{micro}).}
A face detector followed by a multi-task action-unit head emits eight AU intensities per detected
face, $\widehat q_{j,t}\in[0,1]$ for $j\in\mathcal{A}_{\mathrm{FACS}}=\{1,2,4,6,9,12,25,26\}$.
Against the EmoScript target intensities $q^\star_j$ we score agreement by a per-AU
intersection-over-union overlap (rather than an $\ell_p$ kernel, so a target and an observation
are rewarded for occupying the same activation interval), averaged over face frames:
\begin{equation}
    \mathrm{IoU}(q^\star_j,\widehat q_{j,t})=\frac{\min(q^\star_j,\widehat q_{j,t})}{\max(q^\star_j,\widehat q_{j,t})+\epsilon},\quad
    s_{\mathrm{micro}}=\frac{1}{|\mathcal{A}^\star|}\sum_{j\in\mathcal{A}^\star}
    \frac{1}{|\mathcal{F}|}\sum_{t\in\mathcal{F}}\mathrm{IoU}(q^\star_j,\widehat q_{j,t}),
    \label{eq:app_micro}
\end{equation}
with AU compliance and validity
$c_{\mathrm{AU}}=\tfrac{1}{|\mathcal{A}^\star_{\mathrm{req}}|}\sum_{j\in\mathcal{A}^\star_{\mathrm{req}}}\mathbf{1}[\max_{t\in\mathcal{F}}\widehat q_{j,t}>\theta_{\mathrm{AU}}]$
and $u_{\mathrm{micro}}=|\mathcal{F}|/T$. The provider is inactive when no frame yields a face
with AU readings. In our implementation, $\theta_{\mathrm{AU}}=0.3$.

\paragraph{Expression similarity (\emph{clip\_autt}).}
A temporal vision--language encoder embeds a $T_{\mathrm{clip}}$-frame clip into $\mathbf{v}$ and compares it to
$N_{\mathrm{au}}$ action-unit text-prompt embeddings $\{\mathbf{t}_j\}$ by cosine similarity,
mapping the mean similarity from $[-1,1]$ to $[0,1]$:
\begin{equation}
    \bar s_{\mathrm{au}}=\frac{1}{N_{\mathrm{au}}}\sum_{j}\cos(\mathbf{v},\mathbf{t}_j),\qquad
    s_{\mathrm{clip}}=\tfrac12(\bar s_{\mathrm{au}}+1).
    \label{eq:app_clip}
\end{equation}
An auxiliary discrete head additionally yields a negative-affect probability
$p_{\mathrm{neg}}=\operatorname{softmax}(\text{logits})_{\mathrm{neg}}$, from which the scorer
derives $\hat v=1-p_{\mathrm{neg}}$ and $\hat a=\min(1,\,\lambda_a p_{\mathrm{neg}}+b_a)$; the linear
rescale in Eq.~\ref{eq:app_clip} (rather than $\operatorname{clip}$) preserves the ordering on
clips where $\bar s_{\mathrm{au}}<0$. In our implementation,
$T_{\mathrm{clip}}=16$, $N_{\mathrm{au}}=47$, $\lambda_a=1.5$, and $b_a=0.2$.

\subsection{Semantic emotion understanding: Valence--arousal reading and the alignment seed}
\label{app:prim_sem}

\paragraph{Valence--arousal reading (\emph{r1omni}).}
A multimodal emotion-understanding model produces a free-text emotion description that is parsed
to an emotion label and continuous valence--arousal estimates $(\hat v,\hat a)$. Given EmoScript
targets $(v^\star,a^\star)$ the reward supports three modes (default: valence):
\begin{equation}
    s_{\mathrm{r1}}=
    \begin{cases}
      \tfrac12\big[(1-|\hat v-v^\star|)+(1-|\hat a-a^\star|)\big], & \textsc{alignment},\\[1mm]
      \hat v, & \textsc{valence},\\[1mm]
      \tfrac12(\hat v+\hat a), & \textsc{mean}.
    \end{cases}
    \label{eq:app_r1}
\end{equation}

\paragraph{Text-alignment seed (\emph{macro}).}
When a general video--text alignment model is available it blends its text-alignment head
$\mathrm{TA}\in[0,1]$ with AU compliance weighted by face rate
$w_{\mathrm{face}}=\min(\text{face rate},1)$; otherwise a heuristic reads compliance directly:
\begin{equation}
    s_{\mathrm{macro}}=
    \begin{cases}
      \lambda_{\mathrm{TA}}\mathrm{TA}+(1-\lambda_{\mathrm{TA}})c_{\mathrm{AU}}w_{\mathrm{face}}, & \text{VideoAlign reflector},\\[1mm]
      b_{\mathrm{macro}}+(1-b_{\mathrm{macro}})c_{\mathrm{AU}}\pi_{\mathrm{conf}}, & \text{heuristic (active iff face rate}\ge\theta_{\mathrm{macro}}).
    \end{cases}
    \label{eq:app_macro}
\end{equation}

In our implementation, $\lambda_{\mathrm{TA}}=0.6$, $b_{\mathrm{macro}}=0.3$, and
$\theta_{\mathrm{macro}}=0.1$. The confidence $\pi_{\mathrm{conf}}$ belongs to the
heuristic reflector. Because the alignment-seeded branch shares an evaluator with
the reference outcome, its correlation cannot serve as independent emotion validation.

\subsection{Body and temporal evidence (body, temporal)}
\label{app:prim_bodytemp}

The body provider counts each EmoScript body marker $m\in\mathcal{M}^\star$ satisfied by at least
one frame; the temporal provider locates the frame $t^\star$ of peak expected-AU energy and
scores its distance to the EmoScript window $[t_s,t_e]$:
\begin{align}
    s_{\mathrm{body}}&=\frac{1}{|\mathcal{M}^\star|}\sum_{m\in\mathcal{M}^\star}
    \mathbf{1}\big[\exists\,t:\ b_{m,t}=1\big], \label{eq:app_body}\\[1mm]
    t^\star&=\arg\max_{t}\sum_{j\in\mathcal{A}^\star}\widehat q_{j,t},\\
    s_{\mathrm{temporal}}&=
    \begin{cases}
      1, & t_s\le t^\star\le t_e,\\[1mm]
      \max\!(0,\,1-\dfrac{\min(|t^\star-t_s|,|t^\star-t_e|)}{\max(t_e-t_s,\,\delta_\Omega)}), & \text{else.}
    \end{cases}
    \label{eq:app_temporal}
\end{align}
The window stabilizer is $\delta_\Omega=0.1$ in our implementation.
These scores require their corresponding observations; unavailable evidence receives
the neutral fallback $s_{\mathrm{neutral}}$ and validity zero. We use $s_{\mathrm{neutral}}=0.5$.
In the analyzed traces, the micro, body, and temporal channels are constant at $0.5$;
this does not imply that every generated video lacks a face or body.

\subsection{Chromatic atmosphere (chromatic)}
\label{app:prim_chromatic}

A per-frame chromatic signal blends mean saturation $\bar S_t\in[0,1]$ with the
fraction of warm-hue pixels on the standard $[0,179]$ hue scale:
\[
W_t=\overline{\mathbf{1}[h\le\theta_{\mathrm{warm}}^-\ \vee\ h\ge\theta_{\mathrm{warm}}^+]}.
\]
The clip-level score aligns the frame mean to target valence $v^\star$, linearly mapped to $[0,1]$:
\begin{equation}
    c_t=\lambda_{\mathrm{sat}}\bar S_t+(1-\lambda_{\mathrm{sat}})W_t,\qquad
    \bar c=\tfrac1T\textstyle\sum_t c_t,\qquad
    s_{\mathrm{chromatic}}=\operatorname{clip}_{[0,1]}\!(1-|\bar c-v^\star|).
    \label{eq:app_chromatic}
\end{equation}

In our implementation, $\lambda_{\mathrm{sat}}=0.5$ and the warm-hue thresholds are
$\theta_{\mathrm{warm}}^-=30$, $\theta_{\mathrm{warm}}^+=150$ on the stated hue scale.

\subsection{Kinetic intensity (kinetic)}
\label{app:prim_kinetic}

A per-frame kinetic signal normalizes the mean absolute frame difference using saturation scale $\sigma_{\mathrm{kin}}$,
and the clip-level score aligns the frame mean to the target arousal $a^\star$:
\begin{equation}
    k_t=\operatorname{clip}_{[0,1]}\!(\tfrac{1}{\sigma_{\mathrm{kin}}B_I}\,\overline{|I_t-I_{t-1}|}),\qquad
    \bar k=\tfrac1T\textstyle\sum_t k_t,\qquad
    s_{\mathrm{kinetic}}=\operatorname{clip}_{[0,1]}\!(1-|\bar k-a^\star|).
    \label{eq:app_kinetic}
\end{equation}
In our implementation, $\sigma_{\mathrm{kin}}=0.20$ and the pixel range is $B_I=255$. Appearance features describe potential cues; they are not part of the seven-provider memory update.

\subsection{Motion naturalness (motion)}
\label{app:prim_motion}

Dense optical flow $\mathbf{f}_t$ on grayscale, short-side-$L_{\mathrm{flow}}$ frames yields per-frame
magnitude $m_t=\overline{\|\mathbf{f}_t\|}$ and a global (dominant) motion vector
$\mathbf{g}_t=\overline{\mathbf{f}_t}$. Four bounded sub-scores reward smooth flow, low camera
jerk, low warp residual, and low compensated flicker:
\begin{equation}
\begin{aligned}
    r_{\mathrm{flow}}&=\kappa(\overline{|\Delta m_t|};\,\sigma_{\mathrm{flow}}), &
    r_{\mathrm{cam}} &=\kappa(\overline{\|\Delta^2\mathbf{g}_t\|};\,\sigma_{\mathrm{cam}}),\\
    r_{\mathrm{warp}}&=\kappa(\overline{|\,\mathcal{W}(I_{t-1};\mathbf{f}_t)-I_t\,|};\,\sigma_{\mathrm{warp}}), &
    r_{\mathrm{flick}}&=\kappa(\overline{|\,\mathcal{S}(I_{t-1};\mathbf{g}_t)-I_t\,|};\,\sigma_{\mathrm{flick}}),
\end{aligned}
\label{eq:app_motion}
\end{equation}
where $\mathcal{W}$ is flow-warping and $\mathcal{S}$ a global-shift compensation. With base
$r_{\mathrm{base}}=\sum_{j\in\{\mathrm{flow,cam,warp,flick}\}}\lambda_j r_j$
and activity $E=1-\exp(-\bar m/\sigma_E)$, the arousal-conditioned score and confidence are
\begin{align}
    s_{\mathrm{motion}}&=\lambda_M r_{\mathrm{base}}+(1-\lambda_M)(1-|E-a^\star|),\\
    u_{\mathrm{motion}}&=\lambda_U\min\!(1,\tfrac{T-2}{n_U})+(1-\lambda_U)\min\!(1,\tfrac{\operatorname{std}(m_t)}{\sigma_U}),
\end{align}
reverting to $s_{\mathrm{motion}}=r_{\mathrm{base}}$ when no target arousal is supplied. The
motion gate scales linearly with this observation confidence (Appendix~\ref{app:gates}).
The mixture coefficients obey $\sum_j\lambda_j=1$. In our implementation,
$L_{\mathrm{flow}}=160$, $(\sigma_{\mathrm{flow}},\sigma_{\mathrm{cam}},\sigma_{\mathrm{warp}},\sigma_{\mathrm{flick}})
=(0.35,0.65,0.085,0.11)$, and $(\lambda_{\mathrm{flow}},\lambda_{\mathrm{cam}},\lambda_{\mathrm{warp}},\lambda_{\mathrm{flick}})
=(0.34,0.24,0.24,0.18)$. We use $\sigma_E=1.5$, $\lambda_M=0.82$,
$\lambda_U=0.6$, $n_U=8$, and $\sigma_U=0.035$.

\subsection{Score normalization and validity}
\label{app:prim_norm}

The evaluated fusion consumes the providers' bounded scores directly. An optional
running-statistic normalization is not an additional transformation applied to those
scores. Availability and confidence are retained separately from score magnitude:
unavailable evidence receives a neutral fallback but zero effective contribution.
Motion confidence is incorporated once through its channel gate, while an activity
mask handles unavailable motion observations. This convention avoids counting the same
confidence twice. The micro, body, and temporal channels are constant in the analyzed
reference traces; their neutral readings do not establish the absence of visible faces
or bodies in the generated videos.

%% file: sections/appendix_C_gates.tex
\section{Reflect: Instance-Dependent Gate Construction}
\label{app:gates}
Reflect separates training progress, semantic consistency, observation availability,
and the prompt's intended intensity. A factor is neutral for providers to which it does
not apply. Let $t$ denote the video generation model update step, $s_0<s_1$ the phase-transition endpoints,
$\gamma_{\min}$ a phase floor, and $q_f$ the available face-evidence ratio:
\begin{equation}
 h(t)=\operatorname{clip}_{[0,1]}\!\left(\frac{t-s_0}{s_1-s_0}\right),\quad
 b=\max(\gamma_{\min},h(t))g_{\mathrm{conflict}},\quad
 f=\min(1,q_f/\theta_f).
\end{equation}
The semantic-conflict factor attenuates disagreement
$d=|s_{\mathrm{CLIP\text{-}AUTT}}-s_{\mathrm{R1\text{-}Omni}}|$ above a threshold
$\theta_d$, with lower bound $\gamma_d$. Subtlety is detected from prompt cues or
$a^\star\le\theta_a$; it selects restraint factors $\xi_f$, $\xi_t$, and $\xi_e$
for facial/body, temporal, and semantic channels, respectively. Otherwise these
restraint factors equal one. Before applying observation-availability masks, the gates are
\begin{equation}
\begin{aligned}
 g_{\mathrm{macro}}&=b,&
 g_{\mathrm{micro}}=g_{\mathrm{body}}&=bf\xi_f,&
 g_{\mathrm{temporal}}&=bf\xi_t,\\
 g_{\mathrm{CLIP\text{-}AUTT}}=g_{\mathrm{R1\text{-}Omni}}&=b\xi_e,&
 g_{\mathrm{motion}}&=\lambda_g+(1-\lambda_g)c_{\mathrm{motion}}.
\end{aligned}
\end{equation}
Here $c_{\mathrm{motion}}$ is the optical-flow observation confidence. Availability
masking and continuous confidence together instantiate the validity term in
Eq.~\xdocref{eq:gate}; confidence is not multiplied a second time.

\paragraph{Implementation settings.}
In our implementation, the default warmup endpoints are $s_0=40$ and $s_1=100$,
with $\gamma_{\min}=0.1$. Face gating uses $\theta_f=0.35$; semantic conflict uses
$\theta_d=0.20$ and $\gamma_d=0.35$. For subtle prompts, detected using
$\theta_a=0.4$ or prompt cues, $(\xi_f,\xi_t,\xi_e)=(0.85,0.75,0.80)$;
the motion floor is $\lambda_g=0.55$. These are rule-based controls rather than learned
reward-model parameters. Historical exported traces retain their recorded gates;
the default specification alone does not reconstruct a past run's full configuration.

\paragraph{Effect on normalized fusion.}
A factor shared by every active provider cannot by itself redistribute trust.
For $g_k=\gamma h_k$ with $\gamma>0$, active normalization yields
\begin{equation}
 R_{\mathrm{emo}}=\frac{\sum_k w_kh_ks_k}{\sum_k w_kh_k+\epsilon/\gamma}.
\end{equation}
Except for the stabilizer, the common scale cancels. Reflect changes the relative
allocation through provider-specific factors, invalidation, and composition with the
original reward branch. The face-activity proxy used to group exported traces in
Appendix~\ref{app:retrieval} is distinct from the current video's face-evidence ratio.

%% file: sections/appendix_D_memory.tex
\section{Evolve: RewardMemory Construction and Use}
\label{app:retrieval}

\paragraph{Record the reference run (Figure~\ref{fig:walkthrough}A).}
A completed run with Reflective Reward Gating supplies the experience used by RewardMemory.
Each trace keeps the prompt--video context, raw provider scores, gates, applied weights, and VideoAlign outcome separate.
The context describes target affect and available evidence; its face-activity proxy records whether selected scoring branches depart from their neutral defaults, rather than measuring visible-face coverage.
The recorded scores and outcomes provide the basis for comparing providers.

\paragraph{Build the calibration table (Figure~\ref{fig:walkthrough}B).}
The complete reference bank is grouped by target arousal and face-branch activity, with a global fallback for contexts with insufficient support.
Within each group, reliability is the correlation between a provider's raw scores and VideoAlign outcomes, before gating.
Algorithm~\xdocref{alg:jit} compares each provider with the group's average reliability and converts the difference into a capped weight multiplier (Eq.~\xdocref{eq:jit_weight}).
This calibrates reward allocation without fitting a value network or changing any reward model.

\paragraph{Use the memory in the next run (Figure~\ref{fig:walkthrough}C).}
Algorithm~\xdocref{alg:emoreward} looks up the matching multipliers, combines them with the base weights and current-video gates, and normalizes over active providers.
Memory sets the allocation from prior experience; reflection can still suppress unavailable or unsuitable cues.
The table remains fixed throughout training of the video generation model.
Panel~C illustrates this multiply-and-normalize step using recorded scores and illustrative starting weights: it is a controlled replay, not an additional training run.
Reflect-only fusion omits the memory adjustment; Appendix~\ref{app:walkthrough} explains the worked example.

%% file: sections/appendix_E_derivation.tex
\section{Derivation of the Closed-Form KL Update}
\label{app:derivation}
For positive base weights on the supported simplex, consider
\begin{equation}
\max_{w\in\Delta^{K-1}}\sum_k w_kA_k^{\mathrm{rel}}-
\tau\sum_k w_k\log\frac{w_k}{w_k^0}.
\end{equation}
The Lagrangian has stationary condition
$A_k^{\mathrm{rel}}-\tau(\log(w_k/w_k^0)+1)+\lambda=0$.
Normalizing the resulting exponentials gives
$w_k=w_k^0e^{A_k^{\mathrm{rel}}/\tau}/\sum_jw_j^0e^{A_j^{\mathrm{rel}}/\tau}$.
Strict concavity yields a unique optimum. A shared additive baseline in the utilities
cancels from this unclipped solution.

For $\Delta_k=\operatorname{clip}(A_k^{\mathrm{rel}}/\tau,-c,c)$, the same derivation
applies to clipped utilities $\widetilde A_k=\tau\Delta_k$. Since
$e^{-c}\le\sum_jw_j^0e^{\Delta_j}\le e^c$, normalized weights satisfy
\begin{equation}
 e^{-2c}\le\frac{w_k^{\mathrm{RM}}}{w_k^0}\le e^{2c}.
\end{equation}
The cap bounds relative changes; it does not guarantee a one-hot distribution as
$\tau\to0$. Zero base weights remain outside the support.

\paragraph{Scope of the bound and cold start.}
The interval $[e^{-c},e^c]$ bounds the exported, unnormalized multipliers; the normalized
weight ratio has the wider bound above. Clipping the utilities is not equivalent to
solving the original objective subject to a prescribed hard KL constraint. If all
advantages are zero, the update returns $w^0$. A constant provider has $\rho_k=0$, but
its centered advantage is negative only when the mean reliability is positive; constant
scores therefore do not imply unconditional down-weighting. In the supplied traces,
positive-reliability providers make this relative attenuation observable. None of these
algebraic properties proves improved generalization or emotional fidelity.

\paragraph{Group-relative invariance.}
If $R_i'=aR_i+b$ with $a>0$, then $\mu'=a\mu+b$, $\sigma'=a\sigma$, and
\begin{equation}
A_i'=\frac{R_i-\mu}{\sigma+\epsilon/a}.
\end{equation}
Translation cancels exactly. By contrast,
$\operatorname{CV}(R)=\sigma/(|\mu|+\epsilon)$ is not translation invariant.
Consequently, a reward offset can increase relative dispersion without improving
the standardized group advantage. This is an algebraic property, not an experimental
claim about learning efficiency.

\paragraph{RL Objective for Video Generation}
\label{app:policy_objective}
For completeness, we state the clipped GRPO objective underlying the group advantages
in Eq.~\xdocref{eq:grpo_adv} \citep{shao2024deepseekmath}. Let $h_{i,t}$ denote the generation
state and $a_{i,t}$ its sampled transition in candidate trajectory $i$. Define
$\varrho_{i,t}(\theta)=\pi_\theta(a_{i,t}\mid h_{i,t},p)/
\pi_{\theta_{\mathrm{old}}}(a_{i,t}\mid h_{i,t},p)$. With clipping radius
$\varepsilon_{\mathrm{clip}}$ and reference-policy regularization coefficient $\kappa$,
\begin{equation}
\begin{aligned}
\mathcal J_{\mathrm{GRPO}}(\theta)
={}&\mathbb E_{p,\{x_i\}\sim\pi_{\theta_{\mathrm{old}}}}
\left[\frac1G\sum_{i=1}^{G}\frac1{T_i}\sum_{t=1}^{T_i}
\left(\min\!\left\{\varrho_{i,t}A_i,
\operatorname{clip}(\varrho_{i,t},1-\varepsilon_{\mathrm{clip}},1+\varepsilon_{\mathrm{clip}})A_i\right\}
\right.\right.\\[-2pt]
&\hspace{38mm}\left.\left.-\kappa\,D_{\mathrm{KL}}\!\left(
\pi_\theta(\cdot\mid h_{i,t},p)\,\|\,\pi_{\mathrm{ref}}(\cdot\mid h_{i,t},p)\right)\right)\right].
\end{aligned}
\label{eq:policy_grpo}
\end{equation}
Here $T_i$ is the number of sampled transitions, $\pi_{\theta_{\mathrm{old}}}$ is the rollout
policy, and $\pi_{\mathrm{ref}}$ is a fixed reference policy. The scalar reward of the
completed video determines $A_i$; advantages and reward-provider outputs are treated as
constants during the policy update. This expression states the common clipped objective;
individual video RL algorithms specify their transition likelihoods, sampled timesteps,
and regularization. EmoReward changes the scalar reward supplied to the RL algorithm and preserves
those algorithm-specific choices. The policy KL penalty acts on the video generation
model, while the reward-allocation KL term acts on the provider weights.

%% file: sections/appendix_F_metrics.tex
\section{Reward-Gating Metrics and Reproducibility}
\label{app:gating_metrics}
Table~\xdocref{tab:agentic} reports two reward diagnostics (SNR$^*$ and Sel.), three
normalized pillars (EF, GD, GR), and two composites (core, emo). All seven policies
use the same backbone and optimization budget. The exported evaluation contains
160 four-candidate groups for VideoAlign, single-provider, and random routing, and
152 groups for each remaining policy. Thus the available sample counts differ,
even though the nominal training budget is shared. The references identify scoring
models or routing mechanisms, not reproductions of complete agentic systems.

\subsection{Reward Diagnostics}
Let $R_i$ denote the fused training reward, $y_i$ the VideoAlign reference outcome,
and $s_{i,k}$ provider $k$'s score. A group $g$ contains candidates for one prompt.

\paragraph{Reward--outcome agreement (ROA).}
ROA is an auxiliary input to GR rather than a displayed column in Table~\xdocref{tab:agentic}.
It measures within-group pairwise ranking agreement. For outcome-nontied pairs
$\mathcal C_g=\{(i,j):i<j,\ |y_i-y_j|>\epsilon_y\}$, the intended statistic is
\begin{equation}
 \operatorname{Agr}(g)=\frac1{|\mathcal C_g|}
 \sum_{(i,j)\in\mathcal C_g}
 \left[\mathbf1\{(R_i-R_j)(y_i-y_j)>0\}
 +\frac12\mathbf1\{R_i=R_j\}\right].
 \label{eq:roa_diagnostic}
\end{equation}
The reported agreement is $\mathrm{ROA}=\operatorname{Mean}_g\operatorname{Agr}(g)$. The supplied pairwise helper uses $\epsilon_y=10^{-9}$ and returns $1/2$ when no
outcome-nontied pairs exist. However, its treatment of reward ties differs from the
symmetric half-credit convention above. The original aggregation script is unavailable;
we therefore retain the archived ROA values as inputs to GR, without claiming to have
recomputed their tie handling. VideoAlign-only obtains one by construction when its
reward and reference ordering coincide. ROA measures agreement with that evaluator,
not independent human or emotional quality.

\paragraph{Within-group relative dispersion (SNR$^*$).}
With population standard deviation within each group,
\begin{equation}
 D_g=\frac{\operatorname{Std}_{i\in g}(R_i)}
 {|\operatorname{Mean}_{i\in g}(R_i)|+\epsilon_D},\qquad
 \mathrm{SNR}^{*}=\frac1{|\mathcal G|}\sum_{g\in\mathcal G}D_g.
 \label{eq:snrstar}
\end{equation}
The supplied diagnostic implementation uses $\epsilon_D=10^{-6}$. This is a
coefficient of variation, not a signal-to-noise power ratio. A reward offset can
increase it while leaving centered GRPO advantages unchanged; a high value neither
establishes nor is necessary for better policy gradients. We report the archived
per-policy reductions, separately from the aggregate-level checks below.

\paragraph{Reliability selectivity (Sel.).}
Let $\rho_k=\operatorname{Corr}(s_{\cdot,k},y)$ over the policy's evaluated candidates,
using the nine-channel diagnostic pool $\mathcal P$. For $Z=\sum_{k\in\mathcal P}|\rho_k|>0$,
\begin{equation}
 q_k=\frac{|\rho_k|}{Z},\qquad
 \mathrm{Sel.}=1-\frac{-\sum_{k\in\mathcal P}q_k\log q_k}{\log|\mathcal P|},
 \qquad 0\log0=0.
 \label{eq:selectivity_diagnostic}
\end{equation}
The pool includes chromatic and kinetic readouts as well as the seven fusion providers.
Sel. describes concentration of absolute provider--outcome correlations, not applied
fusion weights. It discards correlation signs and does not establish causal provider
usefulness. All seven reported policies have $Z>0$; the auxiliary zero-sample row and
its all-zero-profile convention are excluded. Recomputing Sel. from the exported
correlation profiles agrees with the stored values within $5\times10^{-5}$.

\subsection{Normalized Pillars and Composite Scores}
For the fixed comparison set $\mathcal A$ of seven policies, define
\begin{equation}
 N(v_a)=\frac{v_a-\min_{b\in\mathcal A}v_b}
 {\max_{b\in\mathcal A}v_b-\min_{b\in\mathcal A}v_b}.
 \label{eq:egb_normalization}
\end{equation}
The supplied reducer returns zero for a constant input across policies; none of the
three reported inputs is constant. Normalization is specific to the compared set,
so the resulting scores are not absolute measures transferable across benchmarks.

\paragraph{EF: preference-proxy pillar.}
$\mathrm{EF}_a=N(I_a)$, where $I_a$ is the archived mean ImageReward score
\citep{xu2023imagereward}. EF retains the benchmark's ``Emotion Fidelity'' abbreviation,
but its input is a general visual preference model. We do not interpret it as an
independent measurement of emotion fidelity. The raw means and normalization inputs
are included in the supplement.

\paragraph{GD: relative-dispersion pillar.}
$\mathrm{GD}_a=N(\mathrm{SNR}^{*}_a)$. The benchmark calls this Gating Discriminativeness;
mathematically, it is the normalized coefficient of variation above. Its inclusion
does not turn reward dispersion into a gradient-quality measurement.

\paragraph{GR: agreement-weighted selectivity pillar.}
$\mathrm{GR}_a=N(\mathrm{Sel.}_a\,\mathrm{ROA}_a)$. GR combines correlation-profile
concentration with agreement with the reference evaluator. Both inputs are diagnostic
summaries; the product does not establish an independent reliability ground truth.

\paragraph{core and emo: two specified aggregations.}
The equal-weight composite and preference-emphasized composite are
\begin{equation}
 \mathrm{core}_a=\frac{\mathrm{EF}_a+\mathrm{GD}_a+\mathrm{GR}_a}{3},\qquad
 \mathrm{emo}_a=\lambda_{\mathrm{EF}}\mathrm{EF}_a+
 \lambda_{\mathrm{GD}}\mathrm{GD}_a+\lambda_{\mathrm{GR}}\mathrm{GR}_a.
 \label{eq:egb_composites}
\end{equation}
Here the nonnegative coefficients sum to one; our implementation uses
$(\lambda_{\mathrm{EF}},\lambda_{\mathrm{GD}},\lambda_{\mathrm{GR}})=(0.5,0.25,0.25)$.
We report the predefined composites as descriptive summaries. Min--max normalization
fixes the endpoints for this set of policies, so small raw differences can lead to
large changes in a normalized pillar.

\subsection{Emotion and visual metrics}
\label{app:dea}
\paragraph{Facial reward during training.}
The facial reward provider uses CLIP-AUTT~\citep{zeeshan2026clipautt} with a
CLIP ViT-B/32 backbone and an action-unit (AU) text adapter and temporal classifier
trained on BioVid. It produces AU similarity scores and probabilities for the
BioVid pain and neutral classes.

\paragraph{Emotion metrics at evaluation.}
DEA and TERS use the standard frozen
CLIP-ViT-H-14-laion2B-s32B-b79K checkpoint, trained on the
LAION-2B subset of LAION-5B~\citep{schuhmann2022laionb}. This encoder is held out
from reward computation, RewardMemory, and checkpoint selection. It encodes
24 uniformly sampled frames per video and one text template for each of the
eight Plutchik emotion rings. Let $\mathbf{v}_f$ be the frame-$f$ embedding
and $\mathbf{t}_r$ the text embedding for ring $r$. The frame-averaged image--text
cosine similarity for each of the eight rings is
\begin{equation}
    s_r = \frac{1}{24}\sum_{f=1}^{24}
             \frac{\mathbf{v}_f^\top\mathbf{t}_r}
                  {\lVert\mathbf{v}_f\rVert_2\lVert\mathbf{t}_r\rVert_2}.
\end{equation}
Let $T$ be the prompt's set of target rings. DEA checks whether the largest
$s_r$ in $T$ exceeds the largest score among the corresponding opposite rings
after excluding opposites that are themselves targets. Of the 302 prompts,
261 have a usable opposite. TERS uses all eight ring scores. We sort them in
descending order with ranks starting at one, using the fixed ring order
(joy, trust, fear, surprise, sadness, disgust, anger, anticipation) for ties.
The score for one video is
\begin{equation}
    r^* = \min_{r\in T}\operatorname{rank}_{\mathbf{s}}(r),
    \qquad \mathrm{TERS} = \frac{8-r^*}{7}.
\end{equation}
A score of one means a target ring ranks first. We compute TERS on all 302
prompts and report its mean and SEM across prompts. Table~\ref{tab:dea_robustness}
compares the same configurations under LAION-2B ViT-H/14 and OpenAI ViT-L/14.
For DEA, we report discordant algorithm--prompt pairs descriptively. The
two-sided paired permutation test flips the base and guided labels jointly
for all three RL algorithms within each prompt, accounting for prompt reuse.

\paragraph{Visual metrics.}
For PickScore and CLIPScore, we divide the frame-level logits by $26$ and $30$,
respectively, then average the scaled scores across frames.

\input{sections/tables/dea_robustness}

%% file: sections/tables/dea_robustness.tex
\begin{table}[ht]
\centering\small
\setlength{\tabcolsep}{4pt}
\caption{\textbf{Emotion metrics across frozen encoders.} DEA uses 261 prompts
and TERS uses all 302. The last two rows summarize discordant algorithm--prompt
pairs and prompt-blocked permutation tests for DEA.}
\label{tab:dea_robustness}
\begin{tabular*}{\linewidth}{@{\extracolsep{\fill}}lcccccccc@{}}
\toprule
& \multicolumn{4}{c}{CLIP ViT-H/14} & \multicolumn{4}{c}{CLIP ViT-L/14} \\
\cmidrule(lr){2-5}\cmidrule(lr){6-9}
& \multicolumn{2}{c}{DEA} & \multicolumn{2}{c}{TERS}
& \multicolumn{2}{c}{DEA} & \multicolumn{2}{c}{TERS} \\
\cmidrule(lr){2-3}\cmidrule(lr){4-5}\cmidrule(lr){6-7}\cmidrule(lr){8-9}
RL algorithm & Base & +EmoReward & Base & +EmoReward
             & Base & +EmoReward & Base & +EmoReward \\
\midrule
DanceGRPO & 0.605 & \textbf{0.632} & 0.764 & \textbf{0.782} & 0.625 & \textbf{0.632} & 0.763 & \textbf{0.770} \\
FlowGRPO & 0.621 & \textbf{0.670} & 0.786 & \textbf{0.802} & 0.625 & \textbf{0.659} & 0.782 & \textbf{0.802} \\
SAGE-GRPO & 0.617 & \textbf{0.674} & 0.788 & \textbf{0.819} & 0.613 & \textbf{0.670} & 0.774 & \textbf{0.808} \\
\midrule
Discordant wins & \multicolumn{2}{c}{90/145 (62.1\%)} & \multicolumn{2}{c}{--}
                & \multicolumn{2}{c}{106/186 (57.0\%)} & \multicolumn{2}{c}{--} \\
\bottomrule
\end{tabular*}
\end{table}

%% file: sections/appendix_G_ablations.tex
\section{Extended Evaluation and Ablations}
\label{app:ablation}
\paragraph{Provider-pool construction.}
Figure~\xdocref{fig:provider_pool} presents the main-text visual comparison;
Table~\ref{tab:exp3_toolpool} quantifies the same four reward configurations used to construct the
provider interface. Semantic emotion and motion cues improve some axes while exposing
trade-offs on text alignment and preference. The configurations also differ in fusion
weights and curricula; their comparison is not a single-factor ablation. In the
semantic-cue configuration, the CLIP channel has zero fusion weight and cannot explain
a training gain attributed to an active CLIP reward.

\begin{table}[ht]\centering\small
\caption{\textbf{Provider-pool configurations.} Higher is better. TA denotes text alignment.
The reward configurations use different weights and curricula; missing VQ is not imputed.}
\label{tab:exp3_toolpool}
\begin{tabular}{lrrrrr}\toprule
Configuration & Overall & VQ & MQ & TA & ImageReward \\\midrule
VideoAlign & -1.614 & -0.505 & -0.313 & -0.797 & -0.866 \\
+Semantic emotion & -1.363 & --- & -0.014 & -0.991 & -0.922 \\
+Motion naturalness & -1.453 & -0.505 & 0.075 & -1.024 & -1.293 \\
Balanced tool pool & -1.178 & -0.297 & 0.074 & -0.954 & -1.113 \\\bottomrule
\end{tabular}\end{table}

\paragraph{Ranking agreement.}
For each four-candidate group, let
$\mathcal C=\{(i,j):i<j,\ y_i\ne y_j\}$. We use
\begin{equation}
\label{eq:agr}
\operatorname{Agr}(\mathcal G)=\frac1{|\mathcal C|}\sum_{(i,j)\in\mathcal C}
\left[\mathbb I\{(R_i-R_j)(y_i-y_j)>0\}+\tfrac12\mathbb I\{R_i=R_j\}\right].
\end{equation}
Groups without comparable pairs are excluded. Agreement is averaged over groups;
reward ties receive half credit independently of sample ordering. The analysis contains
160 baseline groups for DanceGRPO and 168 groups for each other algorithm--reward
configuration, totaling 1,504 groups and 6,016 videos. These training candidates are
separate from the 30-prompt fixed-checkpoint evaluation sets.

\paragraph{Memory sensitivity.}
Across 27 temperature, neighborhood-size, and memory-window settings, the memory rule
exceeds fixed weighting at the final window in 23 settings. None has a positive difference
when averaged over all windows. These settings reuse the same traces and are not
independent training seeds. The results motivate sensitivity analysis rather than a claim
of monotonic improvement with accumulating memory.

\input{sections/tables/memory}

%% file: sections/tables/memory.tex
\begin{table}[t]\centering\small
\caption{\textbf{RewardMemory calibration.} Provider reliability and unnormalized multipliers, estimated from 3,232 reflective reward traces ($\tau=0.5$, $c=1$). Macro includes VideoAlign information, so its association is partly shared with the outcome.}
\label{tab:exp3_jit}
\begin{tabular}{lrrrrrrr}\toprule
Provider & Macro & Motion & CLIP & R1-Omni & Micro & Body & Temporal \\\midrule
$\rho_k$ & 0.429 & 0.332 & 0.073 & 0.047 & 0.000 & 0.000 & 0.000 \\
$m_k$ & 1.833 & 1.510 & 0.899 & 0.854 & 0.778 & 0.778 & 0.778 \\
\bottomrule\end{tabular}
\end{table}

%% file: sections/appendix_H_walkthrough.tex
\section{RewardMemory: A Worked Rollout Example}
\label{app:walkthrough}
\begin{figure}[!ht]\centering
\includegraphics[width=\linewidth]{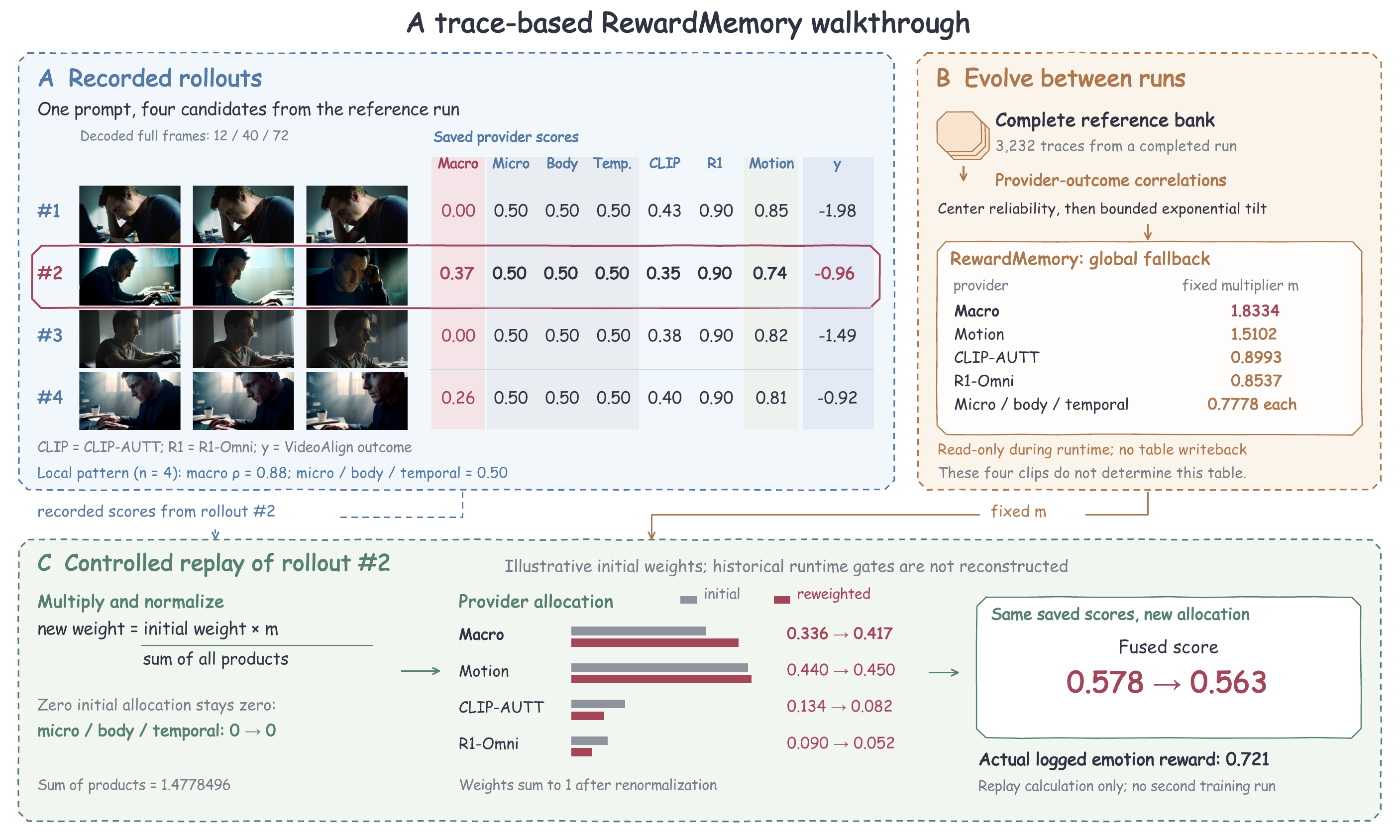}
\caption{\textbf{From completed-run evidence to calibrated fusion.} Recorded observations (A) supply provider scores and outcomes; RewardMemory (B) supplies fixed multipliers estimated from the complete reference bank, rather than the four displayed clips. A controlled replay (C) applies these multipliers to illustrative initial weights for rollout \#2. The resulting $0.578\!\rightarrow\!0.563$ calculation demonstrates reallocation with fixed scores; it is distinct from the logged reward $0.721$ and does not represent a second training run.}
\label{fig:walkthrough}
\end{figure}

Figure~\ref{fig:walkthrough} connects recorded provider scores, a previously exported
calibration table, and a reproducible allocation calculation. Appendix~\ref{app:retrieval} explains the principles behind panels A--C. The clips contain 81 frames at 24 fps and have 624$\times$352 resolution; we display
frames 12, 40, and 72. The four videos are
candidates for one prompt from a reference rollout group. Their scores and outcomes
come directly from the accompanying reward records; the images are decoded full frames.
The macro provider has positive within-group outcome correlation, while three channels
return their neutral constant. These four examples illustrate local behavior, not the
population estimate used to construct RewardMemory.

\paragraph{Separating observations from the replay.}
The exported table's global entry differs from its context-specific entries. We use the
global entry here to isolate the reweighting operation, without claiming that it was
the entry selected for these historical videos. The initial active allocation is
$(0.336,0.440,0.134,0.090)$ over macro, motion, CLIP-AUTT, and R1-Omni,
with zero mass on the other three providers. These are illustrative starting weights,
not reconstructed runtime gates. Multiplying by the exported global multipliers and
renormalizing produces approximately $(0.417,0.450,0.082,0.052)$.
On rollout \#2's recorded scores, the weighted mean changes from $0.578$ to $0.563$;
its actual logged emotion reward is $0.721$. The distinction matters because the
historical reward used its own training phase and configuration. An increase in a
provider's allocation need not increase the scalar fused score, and neither change
alone establishes improved generated quality.

\paragraph{Prompt.}
{\small\itshape The camera slowly pushes forward, a cold side light slanting through the window frame, carving a sharp beam across the dimly lit study and illuminating the man seated at his desk. He leans slightly forward, hands folded over the keyboard, fingers tapping the keys with slow, precise rhythm. Shot from a low angle behind and to the side, the frame captures his furrowed brow and narrowed eyes, his gaze fixed like a needle on the flickering stream of code on the screen. His lips are pressed into a thin line, the jawline taut with concentration, nostrils fluttering faintly with each breath, occasionally biting his lower lip as if swallowing an unspoken question. The light casts a soft shadow across his forehead, making the sharp glint in his eyes even more intense. In the background, scattered papers and empty coffee cups rest silently on the bookshelf, while the rhythmic tapping of the keyboard echoes in the stillness, like the pulse of a heartbeat. He suddenly pauses, raises his right hand to his temple, his eyes briefly drifting, then swiftly refocuses on the screen---a quiet resolve, a steadfast defiance against time.}

%% file: sections/appendix_I_images.tex
\section{Generalization to Image Generation}
\label{app:robust}
We evaluate Stable Diffusion~3.5 and FLUX with three RL algorithms and three reward
configurations on 256 DrawBench prompts per checkpoint: 18 settings and 84 checkpoints.
Internal Emo measures the emotion-stack reward; HPSv3 is a general preference objective
also used to train the baseline. Neither independently measures perceived emotion.

SD3.5 baseline results average three checkpoints at 420--500 updates; calibrated
variants average five at 2,780--3,000. FLUX baseline checkpoints span 2,760--3,000,
and calibrated checkpoints 2,780--3,000. Standard errors describe checkpoint variation,
not independent training seeds. Per-prompt arrays support the reported means;
paired significance requires complete sample and generation-seed mappings and is not claimed.

\paragraph{Observed transfer pattern.}
With SD3.5--DanceGRPO, HPSv3 rises from $9.552$ (baseline) to $10.096$ (Reflect)
and $10.302$ (EmoReward). SD3.5--FlowGRPO declines on both HPSv3 and internal Emo;
FLUX's HPSv3-optimized baseline leads on HPSv3 across RL algorithms. Thus transfer is
objective-dependent rather than uniformly beneficial.

\subsection{Qualitative comparison protocol}
\label{app:repro}
Image plates use four curated prompts per displayed setting; video plates cover three
RL algorithms. Figure~\xdocref{fig:provider_pool} separately compares provider configurations.
These examples illustrate behavior, not a randomly estimated win rate.

Images retain all original pixels. SD3.5 baseline outputs use 500 updates and calibrated
outputs 3,000. The first two FLUX baseline cases use the only saved examples at 500
updates; all other FLUX rows use 3,000. Unequal budgets prevent attributing qualitative
differences solely to calibration. Image seeds are unavailable, so only prompts are matched.
The video rows in Appendix~\ref{app:video_visuals} share prompts and recorded sampling seeds. Clips are 864$\times$480 at 24 fps;
full frames at 0.5, 2.5, and 4.5 seconds are decoded without interpolation or retouching.

The ordinal ratings in Appendix~\ref{app:human} complement these proxy-based and
qualitative comparisons; raw participant/prompt records are needed to quantify uncertainty.
The depicted scenes are generated content.

\input{sections/appendix_images}

%% file: sections/appendix_images.tex
\begin{figure}[!ht]\centering
\includegraphics[width=.86\linewidth,height=.71\textheight,keepaspectratio]{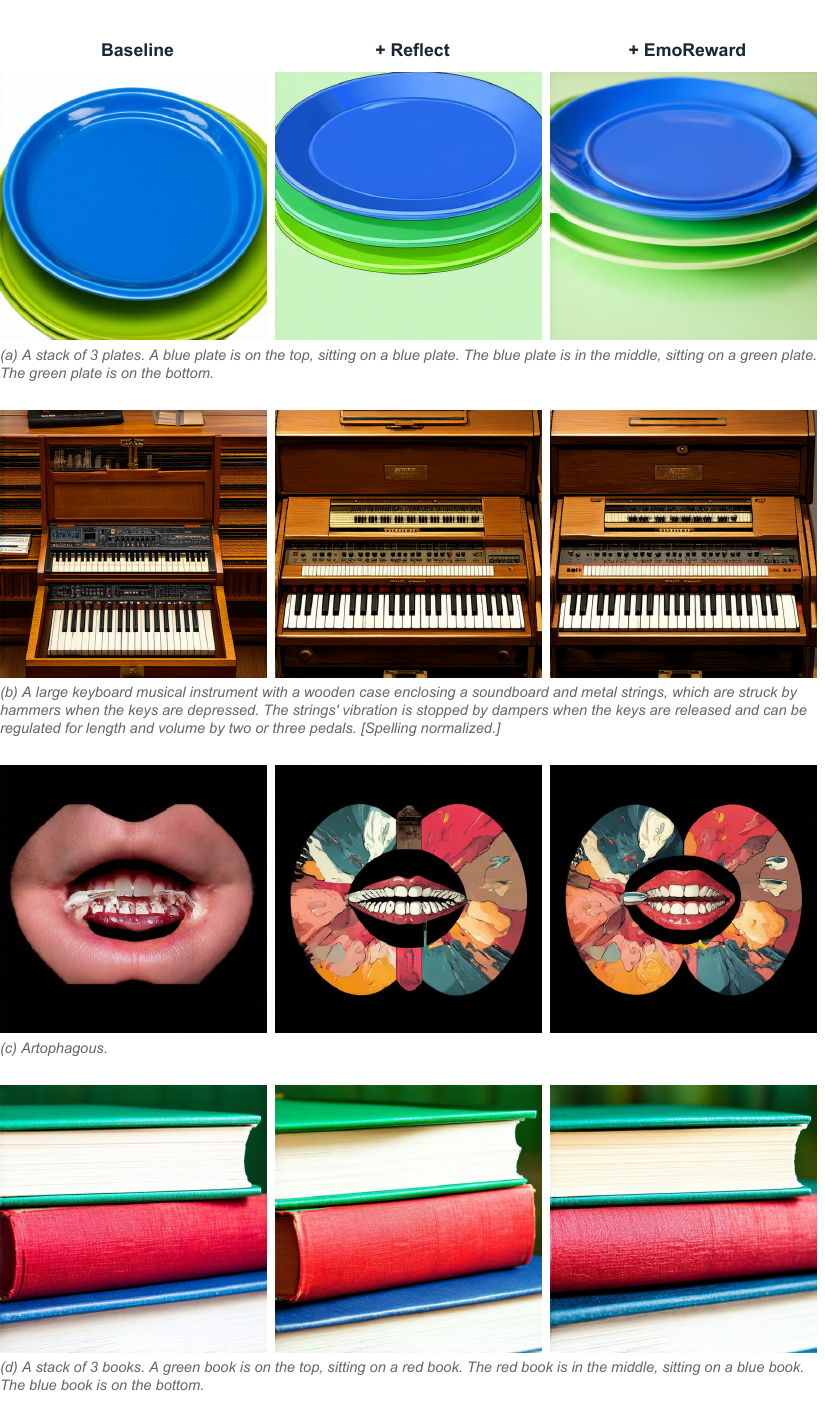}
\caption{\textbf{Qualitative comparison of reward calibration.} Columns show the baseline, the addition of Reflect, and the complete EmoReward configuration for SD3.5 with DanceGRPO. Prompt (b) is spelling-normalized for readability; the images retain the original perturbed input. Budgets and seed limitations appear in Appendix~\ref{app:repro}.}
\label{fig:image-sd35-dance-grpo}
\end{figure}
\begin{figure}[!ht]\centering
\includegraphics[width=\linewidth]{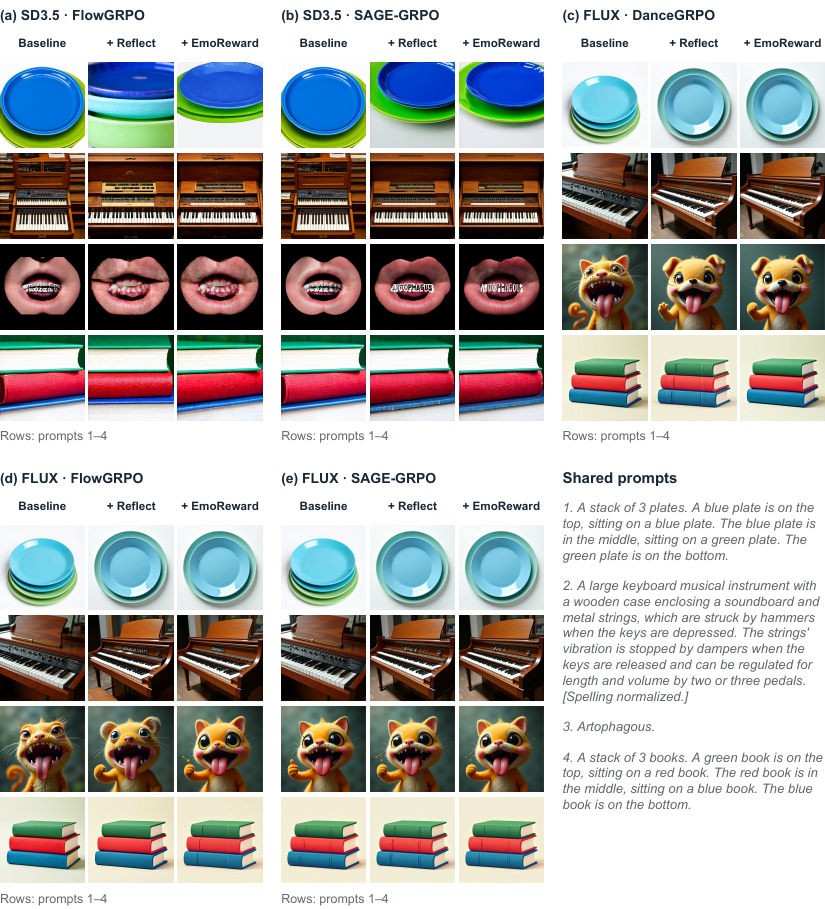}
\caption{Image transfer across SD3.5 and FLUX. Five backbone--algorithm settings share four prompts; each panel compares HPSv3 RL, Reflective gating, and EmoReward. Rows show the same curated prompts as the preceding figure. The first two FLUX baseline rows use earlier saved outputs; see Appendix~\ref{app:repro} for budgets and seed limitations.}
\label{fig:image-generalization-merged}
\end{figure}

%% file: sections/appendix_K_videos.tex
\clearpage
\section{Video Comparisons}
\label{app:video_visuals}
\input{sections/appendix_videos}

%% file: sections/appendix_videos.tex
% Keep each RL algorithm pair on one page in the arXiv page geometry.
\begingroup
% Non-floating captions keep hyperlink targets at each caption.
\captionsetup{hypcap=false}
\setlength{\parskip}{0pt}
\begin{minipage}{\linewidth}
\centering
\includegraphics[width=\linewidth,height=.40\textheight,keepaspectratio]{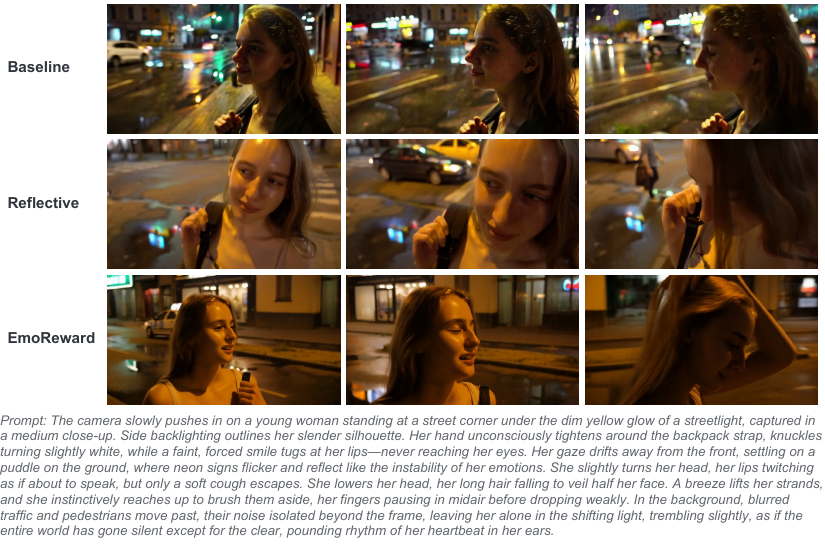}
\captionof{figure}{\textbf{DanceGRPO.} Street-corner hesitation: a strained smile and averted gaze.}
\label{fig:video-dance-020}
\end{minipage}\par
\vspace{14pt}
\begin{minipage}{\linewidth}
\centering
\includegraphics[width=\linewidth,height=.40\textheight,keepaspectratio]{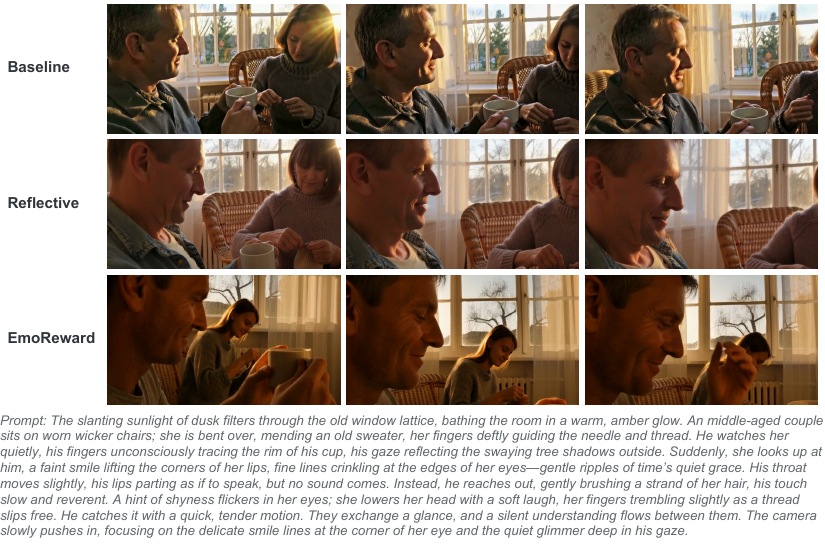}
\captionof{figure}{\textbf{DanceGRPO.} A quiet exchange at dusk: the prompt couples a smile with a tentative hand gesture.}
\label{fig:video-dance-000}
\end{minipage}\par
\clearpage
\begin{minipage}{\linewidth}
\centering
\includegraphics[width=\linewidth,height=.40\textheight,keepaspectratio]{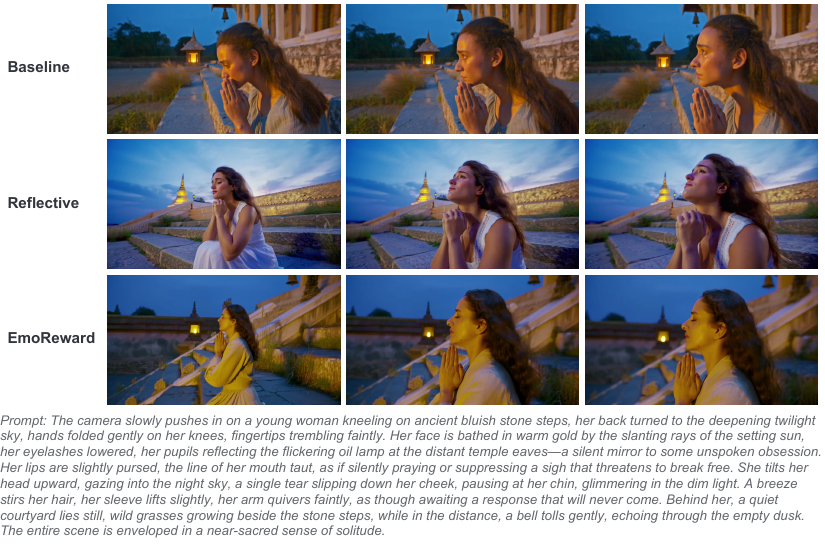}
\captionof{figure}{\textbf{FlowGRPO.} Unease on the stone steps at twilight.}
\label{fig:video-flow-022}
\end{minipage}\par
\vspace{14pt}
\begin{minipage}{\linewidth}
\centering
\includegraphics[width=\linewidth,height=.40\textheight,keepaspectratio]{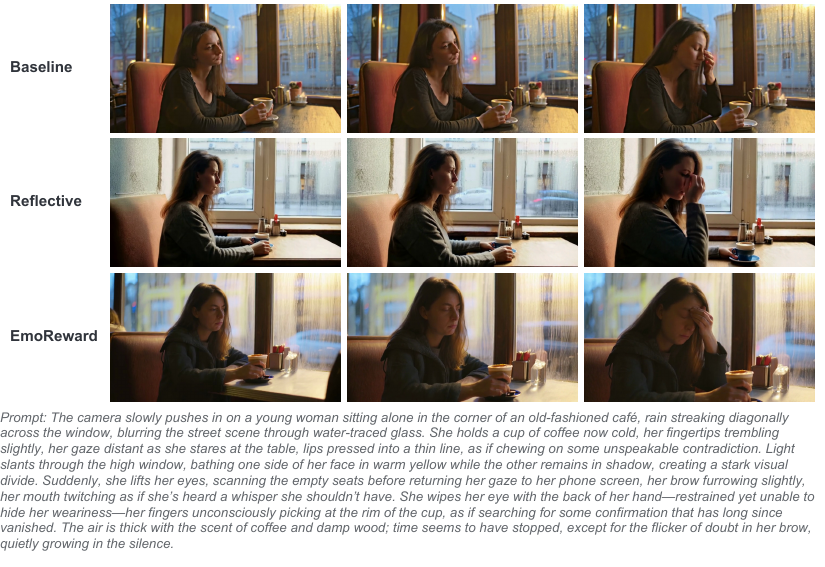}
\captionof{figure}{\textbf{FlowGRPO.} A pause in a rain-streaked caf\'e.}
\label{fig:video-flow-008}
\end{minipage}\par
\clearpage
\begin{minipage}{\linewidth}
\centering
\includegraphics[width=\linewidth,height=.38\textheight,keepaspectratio]{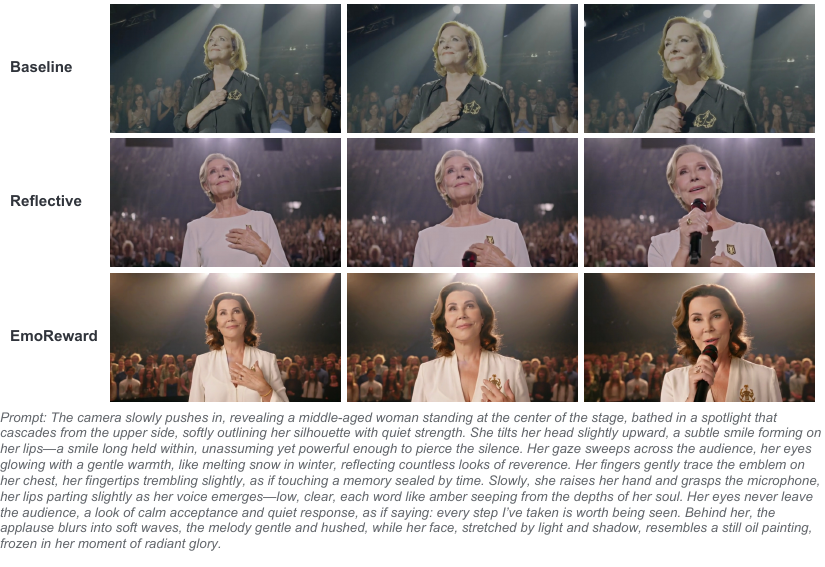}
\captionof{figure}{\textbf{SAGE-GRPO.} Composure under the spotlight in a restrained, emotionally charged stage scene.}
\label{fig:video-sage-027}
\end{minipage}\par
\vspace{14pt}
\begin{minipage}{\linewidth}
\centering
\includegraphics[width=\linewidth,height=.38\textheight,keepaspectratio]{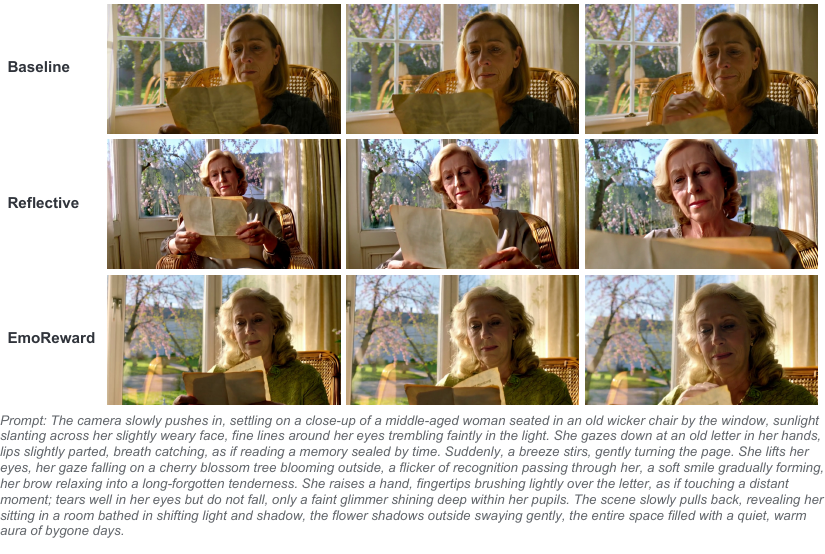}
\captionof{figure}{\textbf{SAGE-GRPO.} Recognition in an old letter in a close-up with gradually softening expression.}
\label{fig:video-sage-029}
\end{minipage}\par
\endgroup